\documentclass[nohyperref]{article}

\usepackage{microtype}
\usepackage{graphicx}
\usepackage{subfigure}
\usepackage{booktabs} 
\usepackage{array}
\newcolumntype{M}[1]{>{\centering\arraybackslash}m{#1}}
\usepackage{float}
\usepackage{multirow}
\usepackage{multicol}
\usepackage{enumitem}
\usepackage{marvosym}

\newcommand{\cross}[1][1pt]{\ooalign{%
  \rule[1ex]{1ex}{#1}\cr
  \hss\rule{#1}{.7em}\hss\cr}}

\usepackage{hyperref}

\usepackage[accepted]{icml2022}

\usepackage{amsmath}
\usepackage{amssymb}
\usepackage{mathtools}
\usepackage{amsthm}

\usepackage[capitalize,noabbrev]{cleveref}

\theoremstyle{plain}

\theoremstyle{definition}

\theoremstyle{remark}

\usepackage[textsize=tiny]{todonotes}

\usepackage{tabularx}

\icmltitlerunning{Tranception: protein fitness prediction with autoregressive
transformers and inference-time retrieval}

\begin{document}

\twocolumn[
\icmltitle{Tranception: protein fitness prediction with autoregressive \\ 
transformers and inference-time retrieval}



\icmlsetsymbol{equal}{\cross[0.4pt]}

\begin{icmlauthorlist}
\icmlauthor{Pascal Notin}{oatml}
\icmlauthor{Mafalda Dias}{marks}
\icmlauthor{Jonathan Frazer}{marks}
\icmlauthor{Javier Marchena-Hurtado}{marks}
\icmlauthor{Aidan Gomez}{oatml,co}
\icmlauthor{Debora S. Marks}{equal,marks}
\icmlauthor{Yarin Gal}{equal,oatml}
\end{icmlauthorlist}

\icmlaffiliation{oatml}{OATML Group, Department of Computer Science, University of Oxford, Oxford, UK}
\icmlaffiliation{marks}{Marks Group, Department of Systems Biology, Harvard Medical School, Boston, MA, USA}
\icmlaffiliation{co}{Cohere, Toronto, Canada}

\icmlcorrespondingauthor{Pascal Notin}{pascal.notin@cs.ox.ac.uk}

\icmlkeywords{Machine Learning, ICML}

\vskip 0.3in
]



\printAffiliationsAndNotice{\cross[0.4pt] Equal senior authorship}

\begin{abstract}
The ability to accurately model the fitness landscape of protein sequences is critical to a wide range of applications, from quantifying the effects of human variants on disease likelihood, to predicting immune-escape mutations in viruses and designing novel biotherapeutic proteins. Deep generative models of protein sequences trained on multiple sequence alignments have been the most successful approaches so far to address these tasks. The performance of these methods is however contingent on the availability of sufficiently deep and diverse alignments for reliable training. Their potential scope is thus limited by the fact many protein families are hard, if not impossible, to align. Large language models trained on massive quantities of non-aligned protein sequences from diverse families address these problems and show potential to eventually bridge the performance gap. We introduce Tranception, a novel transformer architecture leveraging autoregressive predictions and retrieval of homologous sequences at inference to achieve state-of-the-art fitness prediction performance. Given its markedly higher performance on multiple mutants, robustness to shallow alignments and ability to score indels, our approach offers significant gain of scope over existing approaches. To enable more rigorous model testing across a broader range of protein families, we develop ProteinGym -- an extensive set of multiplexed assays of variant effects, substantially increasing both the number and diversity of assays compared to existing benchmarks.
\end{abstract}

\section{Introduction}
\label{Section: Introduction}

\begin{figure*}[t]
\centering
\includegraphics[width=14cm]{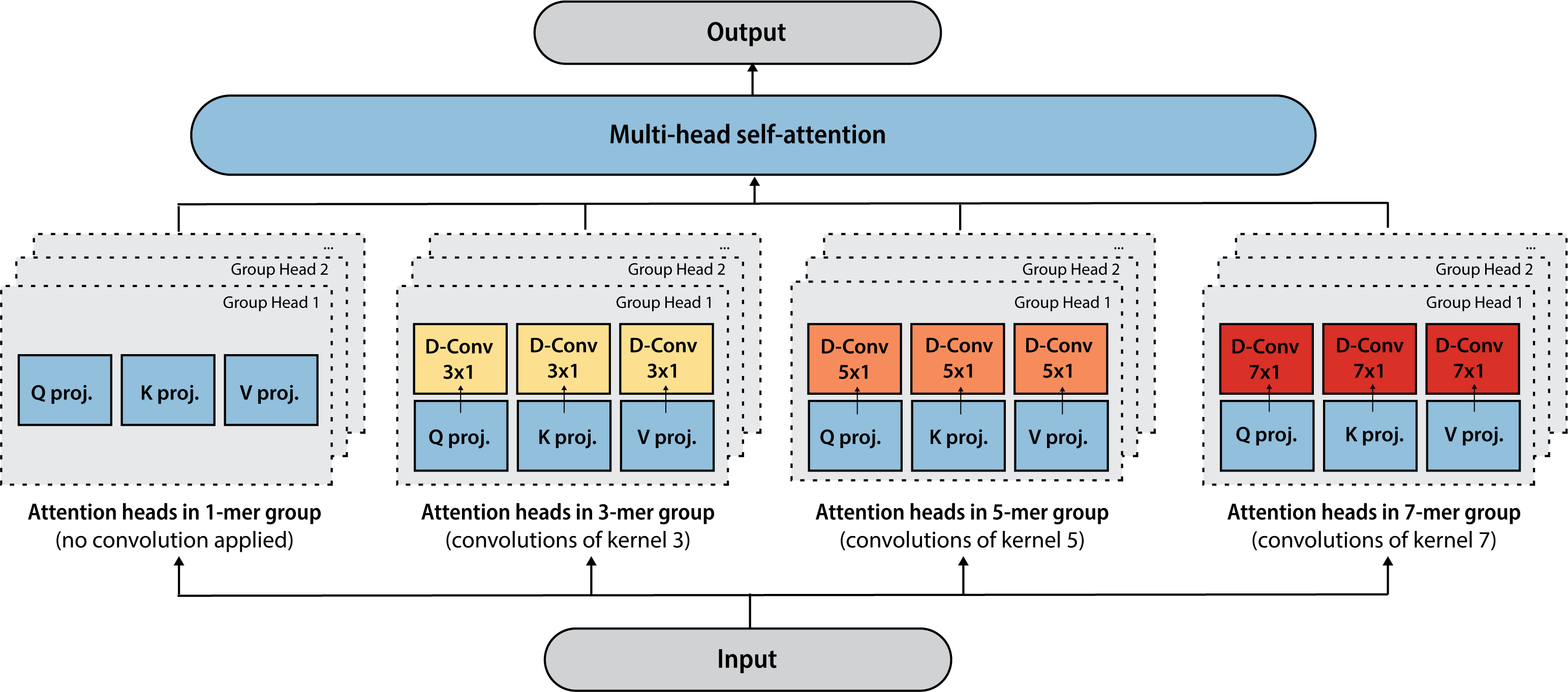}
\caption{\textbf{Tranception attention mechanism.} Attention heads at each layer are split into 4 distinct groups. Except for the first group that does not mix information across tokens (ie., `1-mer' group), separate spatial depthwise convolutions are applied after the Query, Key and Value projections of the other 3 groups with kernel size of 3, 5 and 7 respectively. This incentivizes attention heads within each group to mix information at various ranges across the sequence length. We preserve autoregressiveness of the architecture by applying the right amount of left padding for each convolution.}
\label{Fig: Tranception attention mechanism}
\end{figure*}

Unsupervised models predicting the effects of mutations in protein sequences are emerging as central tools in drug design, pathogen forecasting, identification of disease causing variants and more. Several modeling approaches have been introduced in recent years, offering various trade-offs in terms of performance, diversity of proteins which can be modelled and types of sequence variation which can be scored. 
Current state-of-the-art methods for predicting the effect of single amino acid substitutions are trained on a multiple sequence alignment (MSA) for each protein sequence or domain of interest. In this context, MSAs serve two purposes. First, they act as a data acquisition tool by identifying sequences related to the target in a large protein database, in order to then train a model on a relevant set of sequences. Second, they align sequences by modelling insertions, deletions and substitutions, resulting in a coordinate system which enables the amino acid at a given position to be compared across the training set. 
While training models on protein-specific alignments has been shown to be an effective approach for thousands of proteins \cite{Frazer2021DiseaseVP} it nevertheless brings severe limitations. For instance, such models can not make predictions for sequences which are incompatible with the coordinate system of the MSA used in training (eg., insertions and deletions), thereby limiting scope. Additionally, a large fraction of the proteome corresponds to regions that can not be aligned such as so-called disordered regions -- around half of all human proteins contain regions of at least 40 amino acids classified as disordered \cite{radivojac2004protein, toth2016structured}. 
Even when alignments are accessible, the protein function might be taxa specific, and the MSA algorithm may not retrieve a large enough set of homologous sequences for model training. Alignment-based models may be relatively sensitive to the characteristics of the MSAs they are trained on -- including the choice of hyperparameters used to retrieve these MSAs. Lastly, there is a lack of information sharing across models that are independently trained on different data subsets. 
Recently, language models trained on large quantities of non-aligned \cite{Meier2021.07.09.450648} or aligned \cite{Rao2021.02.12.430858} protein sequences have made first steps towards addressing these issues. 
However, models trained on non-aligned sequences still fail to match the performance of alignment-based methods without further fine-tuning on sequences obtained with a MSA \cite{Meier2021.07.09.450648}. 
Furthermore, approaches based on masked language modeling objectives are unable to estimate the log likelihood of full sequences leading to heuristics when predicting mutation effects -- in particular for multiple mutants, and do not support the scoring of `indels' (Appendix~\ref{Appendix:baselines}). 
To address these limitations, we introduce Tranception, an autoregressive transformer architecture which is pretrained on large quantities of non-aligned sequences and leverages retrieval at inference to achieve state-of-the-art fitness prediction performance.
Our model outperforms all prior baselines, especially on proteins with shallow alignments. Since we do not train on MSAs, our performance is less sensitive to their characteristics (eg., their depth, \S~\ref{Section: Discussion}) and we have the flexibility to operate without them, should the number of sequences retrieved be too low. Lastly, the model has broader scope than any other model to date, while exceeding the performance of more specialized models in particular when extrapolating to mutated proteins that are further away in sequence space.

Our contributions are as follows:
\begin{itemize}
    \item We introduce Tranception, a novel autoregressive transformer architecture that promotes specialization across attention heads for enhanced protein modeling (\S\ref{Section: Tranception});
    \item We combine autoregressive predictions and homology from retrieved sequences at inference (\S~\ref{Section: Retrieval}) to reach state-of-the-art fitness prediction performance on both substitutions and indels (\S~\ref{Section: Results});
    \item We curate an extensive set of multiplexed assays of variant effects -- the ProteinGym benchmarks -- substantially increasing both the number and diversity of assays compared to existing benchmarks \cite{riesselman2018deep} (\S\ref{Section: Benchmark}).
\end{itemize}

\section{Background}
\label{Section: Related work}

\subsection{Mutation effect prediction with aligned sequences}
Predicting the effect of genetic variation using aligned protein sequences from diverse organisms, and in particular, predicting if a variant is likely to be disease-causing in humans, has a long history \citep{ng2001predicting, ramensky2002human, reva2011predicting}. While initial models focused on extracting position-specific information from alignments \cite{ng2001predicting}, subsequent work sought to capture more complex patterns. \citet{hopf2017mutation} proposed to model interactions between pairs of distinct positions with energy based models. \citet{riesselman2018deep} later expanded on the concept with DeepSequence: Variational Autoencoders trained on protein-specific MSAs to learn a distribution of amino acid sequence which capture higher-order interactions. Focusing on predicting the pathogenicity of protein variants in human disease-related genes, EVE \citep{Frazer2021DiseaseVP} subsequently enhanced the DeepSequence architecture to reach higher fitness prediction performance.

\subsection{Modeling proteins without alignments}
\label{subsec:related_works_LM}
While MSAs capture meaningful information about protein functions and structures, they also have certain limitations: not all proteins are alignable and, if they are, the depth of the corresponding alignments may not be enough to train models sufficiently large to learn the complex interactions between residues. This has led to a stream of research investigating alternative modeling approaches that do not rely on aligned sequences. \citet{shin2021protein, weinstein2021structured} developed models that could be trained on non-aligned sequences, although they still relied on MSA routines to recover protein-specific sets of homologous sequences to serve as training data. \citet{alley2019unified, heinzinger2019modeling} were the first to introduce models trained across protein families, relying on LSTM architectures \cite{Hochreiter1997LongSM}. Building on advances in the Natural Language Processing literature to train larger-scale language models, ~\citet{madani2020progen, rives2021biological, nambiar2020transforming} proposed the usage of transformer architectures to model protein sequences. \citet{rao2020transformer} introduced the MSA transformer, an architecture to learn a model of MSAs across thousands of protein families, while ESM-1v \cite{Meier2021.07.09.450648} and ProtTrans \cite{elnaggar2020prottrans} focused on learning patterns exclusively from non-aligned sequences from very large protein databases. Closely related models which combine unsupervised protein embedding with supervised data are seeing diverse applications, from supervised protein design tasks \cite{biswas2021low}, to task agnostic sequence representations \cite{bepler2019learning}, to protein structure prediction \cite{jumper2021highly,Baek2021AccuratePO}.

\subsection{Deep Mutational Scanning benchmarks}

Using a large number of Deep Mutational Scanning experiments (DMS) or Multiplex Assays of Variant Effects (MAVEs) to assess the performance of protein models was first proposed in \citet{hopf2017mutation}, with a benchmark of $\sim$ 20 different assays. \citet{riesselman2018deep} later doubled the size of this benchmark ($\sim$ 40 assays). The list of DMS assays curated to benchmark fitness predictors has seen only modest updates thereafter. More recent benchmarks for protein modeling \cite{DBLP:journals/corr/abs-1906-08230, dallago2021flip} have introduced additional assays focused on assessing model performance across a diverse set of downstream tasks. We include all assays related to fitness prediction from these prior benchmarks when building ProteinGym.

\subsection{Retrieval}
Retrieval aims at identifying objects related to a target one in a reference database to improve the processing or modeling of that object. In the Natural Language Processing literature, retrieval has been leveraged for open domain question answering \citep{Robertson2009ThePR,Wang2018R3RR,Karpukhin2020DensePR} or to augment pretrained language models to find relevant information in massive datasets at inference. \citet{Grave2017UnboundedCM} and \citet{Khandelwal2020GeneralizationTM} both extended language models with a k-NN retrieval over pretrained embeddings at test time for question answering, while ORQA \cite{Lee2019LatentRF} and REALM \cite{Guu2020REALMRL} architectures jointly trained both the `retriever' and `reader' models end-to-end. RAG \cite{Lewis2020RetrievalAugmentedGF} then applied similar concepts to the broader task of generative language modelling. More recently, RETRO \cite{Borgeaud2021ImprovingLM} demonstrated the benefits of retrieval at the scale of trillions of tokens.
The bioinformatics literature has also heavily contributed to retrieval systems, in particular through Multiple Sequence Alignments \cite{Thompson1994CLUSTALWI,Edgar2004MUSCLEMS,Sievers2011FastSG,Remmert2012HHblitsLI} who may be used to both retrieve and align homologous sequences.
\section{Tranception}
\label{Section: Tranception}

Tranception is a novel autoregressive transformer architecture that was designed with two core principles in mind: 1) promoting specialization across attention heads 2) explicitly extracting patterns from contiguous subsequences.

\subsection{Tranception attention}
\begin{figure}[t]
\includegraphics[width=8.5cm]{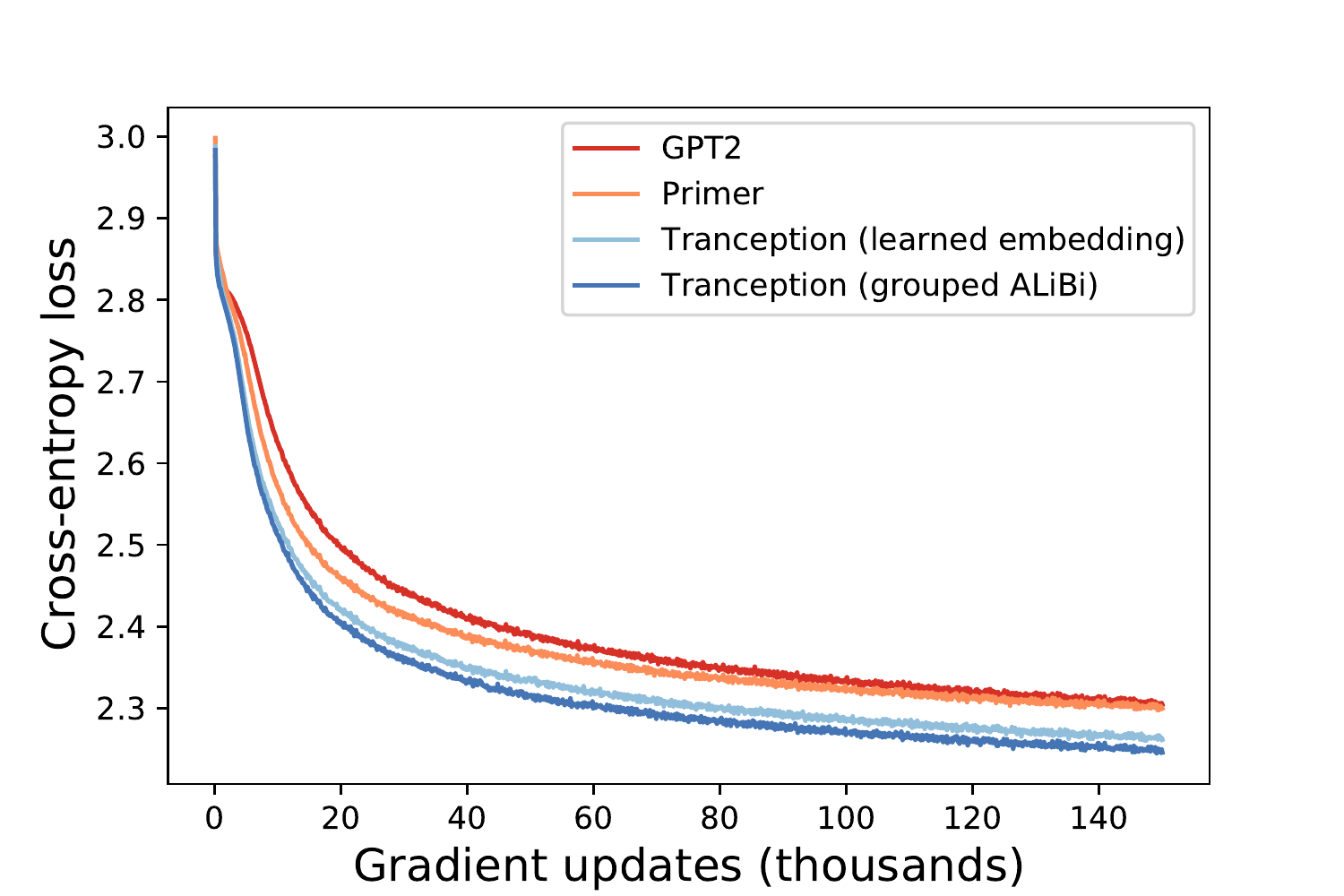}
\caption{\textbf{Training loss comparison across transformer architectures for protein modeling.} We plot the training loss as a function of the number of gradient steps for GPT2 \cite{gpt22018radford}, Primer \cite{so2021primer}, Tranception with learned position embeddings and Tranception with grouped AliBi. All models have similar number of parameters and only differ by they attention mechanism, non-linear activations and position encodings. Tranception converges faster and to a lower loss compared with other architectures. This translates into higher downstream task performance (Appendix~\ref{Appendix: Ablation studies}).}
\label{Fig:Training_loss_GPT2_Primer_Tranception_M}
\end{figure}

The concept of `k-mers' is well-established in biological sequence analysis: k-mers are contiguous subsequences of k elements (typically nucleotides or amino acids) which have proved to be critically useful abstractions in several applications such as de novo assembly, read correction, repeat detection, comparison of genomes, metagenomes \cite{manekar2018benchmark}. The majority of protein language models to date (\S~\ref{subsec:related_works_LM}) have focused on extracting patterns (via sequence tokenization or attention mechanisms) at the amino acid level only. In this work we investigate the benefits from explicitly attending over contiguous subsequences of amino acid tokens via a novel attention mechanism -- Tranception attention (Fig.~\ref{Fig: Tranception attention mechanism}) -- which combines ideas from Primer \cite{so2021primer} and Inception \cite{szegedy2014going} networks. Similar to Primer, we leverage squared ReLU activations and depthwise convolutions after the different multi-head attention projections. Instead of using similar-sized kernels for each depthwise convolution, we split the attention heads at each layer in 4 groups and apply convolutions with different kernel sizes on each group, thereby combining information at different resolutions as in Inception. This incentivizes each attention head to specialize to pattern extractions at different k-mer sizes and leads to both more efficient training and downstream task performance compared with Primer and GPT2 \cite{gpt22018radford} (Fig. \ref{Fig:Training_loss_GPT2_Primer_Tranception_M} and Appendix~\ref{Appendix: Ablation studies}).

\subsection{Grouped ALiBi position encoding}
\label{Section: Grouped ALiBi}
In order to further promote specialization across attention heads and enhance predictions for protein sequences that are longer than the context length, we replace the learned or sinusoidal position encodings typically used in autoregressive transformer architectures \cite{vaswani2017attention, brown2020language}, with a variant of ALiBi \cite{press2021train} called `Grouped ALiBi'. Similar to ALiBi, we remove the position encodings added to input token embeddings, and bias the query-key attention scores with a term proportional to their distance. We however apply the mechanism on each group of attention heads independently, in adequacy with the Tranception attention scheme, to enable each group to learn attention patterns at different distances. Compared to using learned position encodings, this helps reduce the number of parameters, converge faster during training and leads to better downstream task performance (Fig. \ref{Fig:Training_loss_GPT2_Primer_Tranception_M} and Appendix~\ref{Appendix: Ablation studies}).

\subsection{Data processing and augmentations}
Our models are trained on UniRef \cite{10.1093/bioinformatics/btu739}, a large scale protein sequence database. We perform thorough ablations when developing Tranception (Appendix~\ref{Appendix: Ablation studies}). Similar to \citet{Meier2021.07.09.450648}, we investigate the impact of training on protein sequences clustered at different levels of similarity. Unlike what was observed for masked-language model architectures, we find that keeping as much of the granularity available in the dataset is beneficial to downstream task performance. We therefore train our final model (700M parameters) on UniRef100 which, after preprocessing (Appendix~\ref{Appendix: Data processing and augmentations}), leads to a training dataset of $\sim$250 million protein sequences. Our vocabulary is comprised of the standard 20 amino acids \cite{kessel2018intro_proteins}. We find that averaging the predictions obtained by scoring each sequence and its reverse at inference time leads to higher downstream performance (Appendix~\ref{Appendix:scoring_protein_sequences}), and therefore apply sequence mirroring at random during training to teach our model to score sequences from both directions. Our model has a maximum context size of 1024 tokens, which is wider than the length of 98\% of protein sequences in UniRef100 (Table \ref{Appendix - Table: Uniref100 statistics after filtering}). At train time, if a protein is longer than the maximum context size of the model (after accounting for the special start and end of sequence tokens), we extract a randomly-selected contiguous slice of width equal to that maximum context size. Indeterminate amino acids are imputed at random during training and inference. 

\subsection{Scoring sequences for fitness prediction}
\label{Section: Tranception - Scoring sequences}
The goal of fitness prediction is to assess the effects of mutations (eg., amino acid substitutions, insertions or deletions) on the ability of the corresponding mutated protein sequence to perform its function. A common approach to estimating mutation effects is to quantify the likelihood ratio between the mutated sequence and a naturally occurring reference sequence for that protein family, referred to as the `wild-type' sequence \cite{riesselman2018deep}. More formally, we represent each protein $x$ as a sequence of amino acids $(x_1,x_2,...,x_l)$. Our model is trained in a self-supervised fashion to predict the next token $x_i$ in the sequence based on the context of the prior $i-1$ tokens, such that the probability of the full sequence factorizes as:
\begin{equation}
\label{equation:sequence_proba_autoregressive}
    P(x) = \prod_{i=1}^{l}{P(x_i|x_1,...,x_{i-1})} = \prod_{i=1}^{l}{P(x_i|x_{<i})}
\end{equation}

The fitness $F_{x}$ of a given mutated protein $x^{mut}$ is then measured via the log-likelihood ratio with the wild-type sequence $x^{wt}$:
\begin{equation}
\label{equation:log_likelihood_ratio}
    F_{x} = \log \frac{P(x^{mut})}{P(x^{wt})}
\end{equation}

When assessing fitness in sequences longer than the context length, we select the sequence slice providing the widest left and right context for the set of mutations considered (Appendix~\ref{Section: Tranception - Scoring sequences}). At inference time and building on our data augmentations, we take the arithmetic average of the log-likelihood ratios obtained by scoring each sequence and its reverse.
\section{Inference-time retrieval}
\label{Section: Retrieval}

\subsection{Multiple sequence alignments}

Multiple sequence alignments retrieve neighboring proteins in sequence space and align them in the position coordinate system of the seed sequence. The overwhelming majority of fitness prediction models rely on MSAs as they inherently capture critical information about homology, phylogeny, and 3D structure of the corresponding protein family \cite{Thompson1994CLUSTALWI, Thompson1997TheCW}. At a given position, the observed distribution of amino acids over sequences in the MSA recapitulates evolutionary constraints: the protein sequences that are part of the MSA are the variants that maintain fitness and that were not selected out by evolution. 

\subsection{Two modes of inference}

\begin{figure}
\includegraphics[width=8cm]{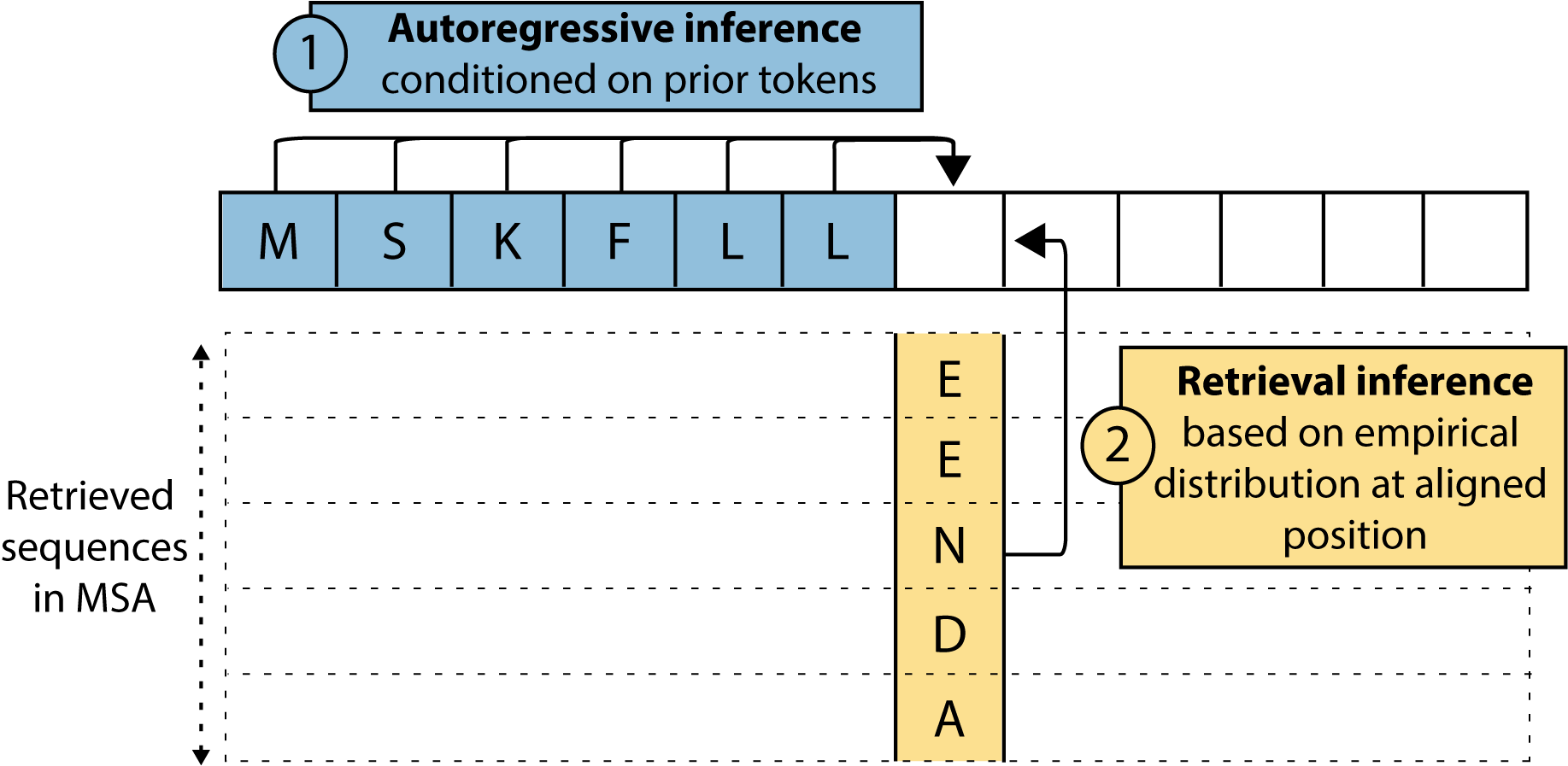}
\caption{\textbf{Combining autoregressive inference and retrieval inference.} Predictions in Tranception are based on two complementary modes of inference: autoregressive predictions based on the context of previously generated tokens and predictions based on the empirical distribution of amino acid at each position in the retrieved set of homologous sequences.}
\label{Fig: AR and MSA combination}
\end{figure}

We propose to augment the autoregressive inference mode from Tranception with a second inference mode -- retrieval inference (Fig.~\ref{Fig: AR and MSA combination}) -- that is based on the empirical distribution of amino acids observed across sequences in the MSA. To that end, the first step consists of retrieving a MSA at inference time for a wild-type sequence from the protein family of interest. When focusing on amino acid substitutions, the retrieved set of homologous sequences is common to both the wild-type and the mutated sequences: we do a retrieval step once per family, and amortize the cost over all mutated sequences to be scored. When dealing with insertions and deletions, we tailor the retrieved MSA to each mutated sequence by deleting columns in the MSA corresponding to deleted positions and adding zero-filled columns in the MSA at inserted positions in the mutated protein. At inference time, the inserted columns or fully non-covered positions are ignored and the model solely relies on its autoregressive mode to make predictions at these positions. 
The second step is to compute the empirical distribution of amino acids for each aligned position based on pseudocounts (ignoring gaps) and Laplace smoothing \cite{Jurafsky2008SpeechAL} (Appendix~\ref{Appendix:retrieval}). Since the distribution of sequences found in protein databases is biased by human sampling (certain organisms are more studied than others) and evolutionary sampling (groups of species that arose from large radiations will have over-represented proteins), we re-weigh sequences in the MSA using the scheme described in \citet{hopf2017mutation}. 
Finally, we estimate the log likelihood $\log P(x)$ for a protein sequence $x$ by a weighted arithmetic average of the log likelihood $\log P_{A}(x)$ from the autoregressive inference mode and the log likelihood $\log P_{R}(x)$ obtained from the retrieval inference mode. This can be equivalently viewed as a weighted geometric average in probability space, and form a proper probability distribution up to a normalization constant:
\begin{equation}
    \log P(x) = \frac{1}{C}[(1-\alpha)\log P_{A}(x) + \alpha\log P_{R}(x)]
\end{equation}
The normalization constant $C$ cancels out when computing the log likelihood ratio as per equation \ref{equation:log_likelihood_ratio}, and thus we do not need to estimate it in practice.
Using equation \ref{equation:sequence_proba_autoregressive}, we further obtain:
\begin{equation}
\label{equation:log_proba_two_modes}
    \log P(x) \propto \sum_{i=1}^{l}{[ (1-\alpha) \log P_{A}(x_i|x_{<i}) + \alpha\log P_{R}(x_i)]}
\end{equation}

The above scoring decomposition as the sum of position level scores is advantageous in practice as it allows us to ignore one of the two inference modes as needed (eg., ignoring retrieval inference at positions with insertions when scoring indels).
Following the scoring procedure described in \S~\ref{Section: Tranception - Scoring sequences}, we first traverse the sequence in the canonical order from left to right and compute the sequence log probability as per equation\ref{equation:log_proba_two_modes} (in practice, all computations are performed in parallel across positions using masks in the self-attention layers to preserve autoregressiveness). We then perform the symmetric operation traversing the protein sequence from right to left, and finally average the two log likelihood ratios.
Retrieving the MSA and computing the corresponding pseudocounts at inference is relatively cheap computationally. Since we only rely on aggregate statistics per position, our proposed approach is not very sensitive to the characteristics of the MSA for the protein family of interest (Fig.~\ref{Fig: filtering sensitivity}), nor to the hyperparameters chosen to retrieve that MSA. As Tranception is trained on a diverse set of non-aligned sequences it is not subject to the biases that may stem from solely training on proteins that can be aligned, and thus exhibits higher performance on difficult to align sequences such as disordered proteins (\S~\ref{Section: Discussion}).

\section{ProteinGym}
\label{Section: Benchmark}

ProteinGym is an extensive set of Deep Mutational Scanning (DMS) assays (Appendix~\ref{glossary}) curated to enable thorough comparisons of various mutation effect predictors in different regimes. ProteinGym is comprised of two benchmarks: 1) a \emph{substitution benchmark} which consists of the experimental characterisation of $\sim$1.5M missense variants across 87 DMS assays 2) an \emph{indel benchmark} that includes $\sim$300k mutants across 7 DMS assays.

The relationship between protein fitness measured experimentally and as reflected by the distribution of sequences selected by evolution is complex. The fitness landscape of naturally occurring proteins is the result of an intricate set of overlapping constraints that these proteins are subjected to in an organism. Consequently, it is often challenging to identify a single molecular property that is both easy to measure experimentally and that reflects that complexity.
To build this benchmark we therefore prioritized assays where both the experimentally-measured property for each mutated protein is expected to reflect the role of the protein in organism fitness as well as - where available - their quality measured via experimental replicates.
The resulting set of DMS assays covers a wide range of functional properties (eg., thermostability, ligand binding, aggregation, viral replication, drug resistance), spans a diverse protein families (eg., kinases, ion channel proteins, g-protein coupled receptors, polymerases, transcription factors, tumor suppressors) and different taxa (eg., humans, other eukaryotes, prokaryotes, viruses). 

ProteinGym is the largest and most diverse set of DMS experiments specifically targeted at the task of variant effect prediction. It contains more than twice the number of assays and variants present in the DeepSequence benchmark \cite{riesselman2018deep}, which had been created for the same purpose (Table \ref{Table: ProteinGym}). 
While most of the curated DMS assays (in our benchmark or others) probe the effect of single amino-acid substitutions, our collection also includes several multiple amino-acid variants which are critical to assess the ability of models to extrapolate further away in sequence space from the naturally occurring proteins they are trained on.
Lastly, as most mutation effect predictors are not able to quantify the effect of insertions and deletions, indels have been absent from the majority of prior benchmarks. We expand on the set of DMS available in \citet{shin2021protein} and \citet{dallago2021flip} to address this gap.

The relationship between protein function and organism fitness has been shown to often be non-linear \cite{boucher2016quantifying}. As such, we use Spearman's rank correlation coefficient between model scores and the experimental measurements as the standard measure of model performance \cite{riesselman2018deep, Meier2021.07.09.450648}. In certain instances, the DMS measurements are characterized by a bimodal profile for which rank correlations are not well suited. To that end, we provide additional measures of model performance: the area under the ROC curve (AUC) and Matthews correlation coefficient (MCC) between model scores and the experimental measurements (Appendix~\ref{Appendix:performance_reporting}).

\begin {table*}[t]

\begin{center}

\begin{tabular*}{\textwidth}{ @{\extracolsep{\fill}} llcccc}

\toprule
\textbf{Measure}  & \textbf{Category} & \textbf{DeepSequence} & \textbf{ProteinGym} & \textbf{Fold increase} 
\\
  \midrule
 \multirow{5}{*}{\begin{tabular}[c]{@{}l@{}} Number of assays \\ by taxon\end{tabular}} & Human & 9 &	33 & 3.7 
 \\
 & Other eukaryotes & 10 &	14 & 1.4 
\\
 & Prokaryotes &  13 & 24 &	1.8 
\\
 & Virus & 5 &	22 & 4.4 
 \\
  & \textbf{All taxa} & \textbf{37} & \textbf{93} &	\textbf{2.5} 
  \\
  \midrule
   \multirow{4}{*}{\begin{tabular}[c]{@{}l@{}} Number of variants \\ by type \end{tabular}} & Single substitutions & 0.12M & 0.36M & 2.9  
 \\
  & Multiple substitutions & 0.55M & 1.26M & 2.3
 \\
   & Indels & 0 & 0.27M & - 
  \\
 & \textbf{All variants} & \textbf{0.67M} &	\textbf{1.89M} &	\textbf{2.8} \\
\bottomrule
\end{tabular*}

\label{Results - Table: Main results}
\end{center}
\caption{\textbf{Comparison of the ProteinGym and DeepSequence benchmarks}. ProteinGym contains a substantially higher number of assays and variants compared to DeepSequence. It addresses notable gaps such as the limited number of viral DMS assays, limited multiple substitutions assays and absence of indels benchmark.}
\label{Table: ProteinGym}
\end{table*}

\section{Results}
\label{Section: Results}
\begin {table*}[t]

\begin{center}

\begin{tabular*}{\textwidth}{ @{\extracolsep{\fill}} llccccc}

\toprule
\textbf{Model}  & \textbf{Model} & \multicolumn{4}{c}{\textbf{Spearman's rank correlation by MSA depth $\uparrow$}} & \textbf{AUC $\uparrow$}\\
\textbf{type} & \textbf{name} &  \textbf{Low}  & \textbf{Medium} & \textbf{High} & \textbf{All} & \textbf{All} \\
\toprule
 \multirow{5}{*}{\begin{tabular}[c]{@{}l@{}}Alignment- \\ based \\ models\end{tabular}} & Site indep & 0.428 & 0.403 &	0.350 &	0.397 & 0.725 \\
 & Wavenet & 0.319 & 0.398 & 0.469 & 0.398 & 0.725 \\
 & DeepSequence & 0.375 & 0.397 & 0.506 & 0.415 & 0.733 \\ 
 & EVmutation & 0.401 &	0.421 &	0.468 &	0.427 & 0.738 \\
 & EVE & 0.408 &	\textbf{0.440} &	\textbf{0.507} &	0.448 &  0.751\\

\midrule
\multirow{4}{*}{\begin{tabular}[c]{@{}l@{}}Protein \\ language \\ models\end{tabular}} & ESM-1v & 0.321 &	0.348 &	0.484 &	0.371 & 0.713 \\
 & MSA Transformer & 0.373 &	0.418 &	0.482 &	0.422 & 0.737 \\ 
  & Tranception (w/o retrieval) & 0.394 & 0.398 & 0.439 & 0.406 & 0.728 \\
  & Tranception (w/ retrieval) & \textbf{0.453} &	0.438 &	0.488 &	\textbf{0.451} & \textbf{0.754} \\
\bottomrule
\end{tabular*}

\end{center}
\caption{\textbf{Average AUC and Spearman's rank correlation between model scores and experimental measurements by MSA depth on the ProteinGym substitution benchmark.} Alignment depth for the various proteins is measured by the ratio of the effective number of sequences $N_{\rm eff}$ in the MSA, using the same weighting scheme as in \cite{hopf2017mutation}, by the length covered $L$: shallow alignments (ie., `Low' group) typically have a low value of $N_{\rm eff}/L$. Specifically, Low: $N_{\rm eff}/L$ $<$1; Medium: 1$<$ $N_{\rm eff}/L$ $<$100; High: $N_{\rm eff}/L$ $>$100. Tranception outperforms all other baselines overall, with the largest performance gaps observed on the low depth proteins.}
\label{Results - Table: Performance by MSA depth}
\end{table*}
\begin {table*}[t]

\begin{center}

\begin{tabular*}{\textwidth}{ @{\extracolsep{\fill}} llcccccc}

\toprule
\textbf{Model}  & \textbf{Model} & \multicolumn{6}{c}{\textbf{Spearman's rank correlation by mutation depth $\uparrow$}} \\
\textbf{type} & \textbf{name} &  \textbf{1}  & \textbf{2} & \textbf{3} & \textbf{4} & \textbf{5+} & \textbf{All} \\
\toprule
\multirow{5}{*}{\begin{tabular}[c]{@{}l@{}}Alignment- \\ based \\ models\end{tabular}} & Site indep & 0.396 & 0.325 & 0.286 & 0.319 & 0.421 & 0.397 
\\
 & Wavenet & 0.394 & 0.344 & 0.329 &	0.281 &	0.396 &	0.398 \\
 & DeepSequence & 0.415 & 0.394 & 0.372 & 0.304 & 0.418 & 0.415
 \\
 & EVmutation & 0.427 & 0.392 & 0.379 & 0.319 & 0.433 & 0.427
 \\
 & EVE & \textbf{0.448} & 0.392 &	0.375 &	0.334 &	0.420 &	0.448 \\

\midrule
\multirow{4}{*}{\begin{tabular}[c]{@{}l@{}}Protein \\ language \\ models\end{tabular}} & ESM-1v & 0.372 &	0.291 &	0.190 &	0.160 &	0.245 &	0.371 \\
  & MSA Transformer & 0.423 &	0.359 &	0.390 &	0.327 &	0.431 &	0.422 \\
 & Tranception (w/o retrieval) & 0.397 &	0.412 &	0.425 &	0.335 &	0.479 &	0.406 \\
  & Tranception (w/ retrieval) & \textbf{0.448} &	\textbf{0.435} &	\textbf{0.443} &	\textbf{0.368} &	\textbf{0.499} &	\textbf{0.451} \\
\bottomrule
\end{tabular*}

\end{center}
\caption{\textbf{Average Spearman's rank correlation between model scores and experimental measurements by mutation depth.}} Mutation depth is measured by the number of distinct substitutions compared with the wild-type sequence.
\label{Results - Table: Performance by mutation depth}
\end{table*}
\begin{table}

\begin{tabular} {lcc}
\toprule
\textbf{Model name} & \textbf{Spearman $\uparrow$} & \textbf{AUC $\uparrow$}\\
\midrule
 Wavenet & 0.412 & 0.724 \\
 Tranception (w/o retrieval) & 0.430 & 0.740 \\
 Tranception (w/ retrieval) & \textbf{0.463} & \textbf{0.759} \\

\bottomrule

\end{tabular}

\caption{\textbf{Average AUC and Spearman's rank correlation between model scores and experimental measurements on the ProteinGym indel benchmark}.}

\label{Results - Table: Performance in indels}
\end{table}

\subsection{Baselines}
\label{subsec:baselines}
We compare the ability of various models to predict the effects of mutations across the DMS assays in ProteinGym. We focus on the main approaches described in \S~\ref{Section: Related work}, including a number of protein-specific alignment-based methods -- Site independent model and EVmutation \cite{hopf2017mutation}, DeepSequence \cite{riesselman2018deep}, EVE \cite{Frazer2021DiseaseVP} -- and large-scale protein language models trained across protein families that leverage alignments during training, such as the MSA Transformer \cite{Rao2021.02.12.430858} or that are alignment-free, such as ESM-1v \cite{Meier2021.07.09.450648}. 
Although technically trained on subsets of unaligned sequences, Wavenet \citep{shin2021protein} models are protein-specific and use MSAs to extract their training data (as such, we group them with other alignment-based models). 
Tranception falls in the category of large-scale protein language models trained across families and, thanks to its two modes of inference, can be seen as a hybrid between ESM-1v and the MSA transformer. As in ESM-1v, it is trained on a large set of unaligned sequences, which makes the training procedure scalable and removes biases that would result from training on alignable proteins only. Similar to the MSA transformer, it leverages the information in a retrieved MSA to enhance fitness predictions. The critical difference is that Tranception is never trained on MSAs and is therefore less sensitive to their characteristics and limitations (Fig.~\ref{Fig: filtering sensitivity}). 
To allow fair comparisons across models we focus on single seed scoring only, but provide additional results with model ensembles in Appendix~\ref{Appendix: Ensembling}.

\subsection{ProteinGym substitution benchmark}
\label{Section: Results - performance comparison}
We compute the Spearman's rank correlation coefficient $\rho$, AUC and MCC between model scores and experimental measurements for all DMS assays in the ProteinGym substitution benchmark (Appendix~\ref{Appendix:detailed_results}, Fig.~\ref{Fig. all_the_spearmans}), and draw very similar conclusions across metrics. 
Tranception with retrieval outperforms all other baselines on the overall benchmark, with markedly higher performance in the regime of low-depth MSAs (Table~\ref{Results - Table: Performance by MSA depth}) and on multiple mutants (Table~\ref{Results - Table: Performance by mutation depth}). When analyzing performance at the taxon level (Table \ref{Appendix_table:spearman_taxa_table}), we observe consistently high performance from Tranception across categories, in particular on human proteins and other eukaryotes. This high performance on human proteins has immediate clinical applications, since Tranception outperforms EVE and directly extends to modelling the entire human proteome, while EVE models need to be trained for each new protein of interest and are available for only $\sim$3k proteins at the time of writing.
Without retrieval, Tranception outperforms ESM-1v, the only baseline which also does not leverage alignments for inference. The performance lift is particularly significant on proteins with shallow alignments, on multiple mutations and viral proteins.

\subsection{ProteinGym indel benchmark}
We report the performance metrics (Spearman, AUC, MCC) on the ProteinGym indel benchmark (Table \ref{Results - Table: Performance in indels}) and compare the performance of Tranception with Wavenet, the only baseline from the set described in \S~\ref{subsec:baselines} that is able to quantify the effects of deletions or insertions. Other alignment-based models are constrained by the fixed coordinate system from the original MSA they have been trained on. ESM-1v and MSA transformer both rely on a scoring heuristic (Appendix\ref{Appendix:baselines}) that requires the mutated position to exist in the wild-type sequence. On the indel benchmark, Tranception outperforms Wavenet both with and without retrieval.

\section{Discussion}
\label{Section: Discussion}

\begin{figure*}[t]
\begin{minipage}[b]{.5\textwidth}
\centering
\includegraphics[width=1\textwidth]{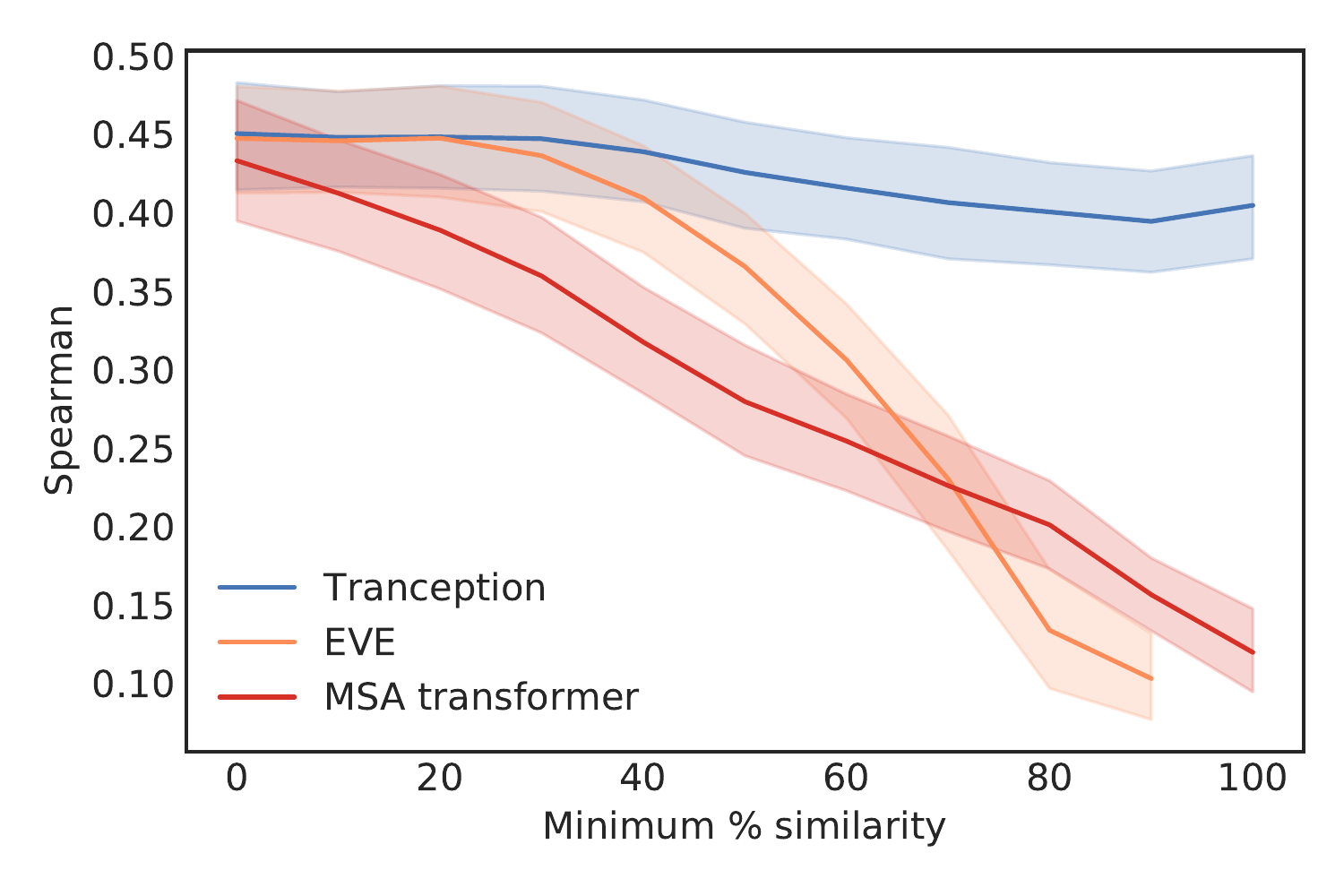}
\end{minipage}
\hfill
\begin{minipage}[b]{.5\textwidth}
\centering
\includegraphics[width=1\textwidth]{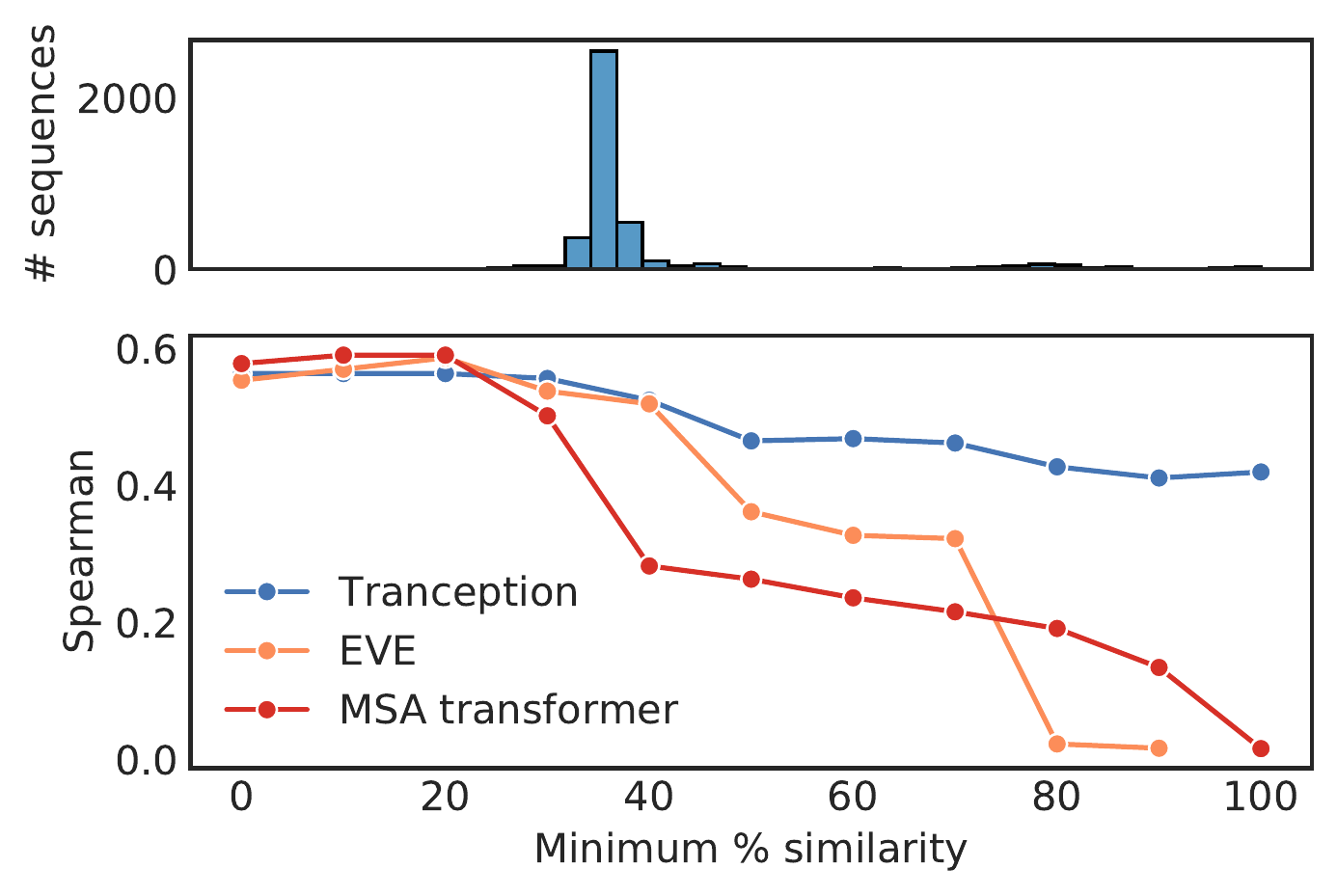}
\end{minipage}

\caption{\textbf{Robustness to alignment depth.} We measure the average performance (Spearman's correlation with DMS measurements) of Tranception, EVE and MSA Transformer as we filter out an increasing proportion of MSA sequences based on their similarity to the seed sequence. The left figure aggregates results across all 87 substitution assays in ProteinGym, the right figure focuses on the tumor protein P53 (\citet{kotler_systematic_2018} assay; other examples are provided in Appendix~\ref{Appendix:filtering_analysis}). The performance of Tranception is robust to MSA depth, while that of EVE and MSA transformer drops significantly as diversity in the MSA is reduced. Rightmost points for Tranception and MSA transformer correspond to performance with no retrieval and with a single-sequence input MSA (the seed) respectively. On the right figure, the histogram reports the number of sequences in the MSA per similarity to the seed sequence groupings.}
\label{Fig: filtering sensitivity}
\end{figure*}

\paragraph{Gain of scope for proteins with shallow alignments.}
Tranception with retrieval outperforms other mutation effects predictors on the ProteinGym substitutions and indels benchmarks.
While retrieval is key to achieve such results, our method only relies on high level statistics of the retrieved alignments. It is thus fairly robust to the various hyperparameters selected to generate these alignments -- even when they result in relatively shallow alignments. When progressively removing sequences in the MSA based on their minimum similarity to the seed sequence (Fig.~\ref{Fig: filtering sensitivity}), we observe that Tranception has consistently high performance, unlike EVE or MSA Transformer.
Given that one can obtain shallow MSAs for far more proteins than deep MSAs, our lightweight inference-time retrieval can be effectively leveraged to enhance mutation effect predictions for the vast majority of proteins.
Disordered proteins are examples of proteins that are notoriously difficult to align. The ProteinGym substitution benchmark includes two proteins with disordered regions, A4 and GCN4. Tranception markedly outperforms all baselines on these two assays (Fig.~\ref{Fig. all_the_spearmans}).
Finally, our approach benefits from the additional flexibility to fully ignore the retrieval-based mode of inference when scoring positions that are not present in the MSAs (eg., indels) and proteins that are not alignable or for which the corresponding set of homologous sequences is extremely shallow (eg., a few sequences).

\paragraph{Gain of coverage. }
A major benefit from Tranception is gain of coverage. For example, BRCA1, the gene encoding Breast cancer type 1 susceptibility protein, has 1863 amino acids making it challenging to model with alignment based methods. Trying to obtain an alignment for the full protein results in poor diversity. Not only alignment-based models (eg., EVE) obtain relatively weak performance when trained on such alignments (Table \ref{Appendix - Table: BRCA1 drill}), but they are unable to make predictions for all mutations impacting regions of the protein with insufficient coverage (see Appendix~\ref{Appendix:baselines} for more details on these limitations). 
We know from previous work~\cite{Frazer2021DiseaseVP} that it is possible to obtain higher quality alignments and model performance when dealing with the two RING and BRCT domains of BRCA1 separately. But this comes at the cost of not being able to make predictions for the majority of the protein (outside of these two domains), and ignore potential dependencies across domains.
Tranception is not subject to these coverage limitations and is able to score all potential mutations of the protein. It obtains relatively high performance without retrieval, yet benefits from leveraging the retrieval inference mode based on full-protein alignment or domain-specific alignments, with slightly higher performance for the latter.

\paragraph{Extrapolating far away in sequence space} Compared with other baselines, Tranception obtains much higher performance on multiple mutants, with the gap widening with mutation depth. While we sought to include as many assays with multiple mutants as possible when building ProteinGym, our benchmarks are still relatively biased towards single mutant assays: on the substitution benchmark, only 11 assays out of 87 included multiple mutants. Since the actual number of multiple mutants is exponentially higher than the number of single mutants for any protein family, it is likely that the reported gap between Tranception and other baselines would only increase on a benchmark with a higher proportion of multiple mutant assays.
Performing well on multiple mutants has important ramifications in several applications, in particular in machine-learning guided protein design for which accurate extrapolation of sequence likelihood far away in sequence space is critical to uncover novel desirable candidate proteins.
\section{Conclusion}

Models of the effects of genetic variations are emerging as powerful tools with diverse applications -- from quantifying genetic predispositions to certain pathologies to designing novel proteins. Yet our understanding of how to build these models is still in its infancy. 
To make progress, we need methods that can scale and perform well across the whole protein universe, including regions that are hard to align or that are recent in evolutionary terms. The model we present in this work makes progress in this direction on five main aspects.

Firstly, the combination of the Tranception model architecture together with retrieval at inference delivers state-of-the-art fitness prediction performance, with a significantly stronger ability to extrapolate to multiple mutants. Scaling up the size of our transformer \cite{Hesslow2022RITAAS}, together with training on a larger and more diverse set of protein sequences \cite{Mitchell2020MGnifyTM,Steinegger2018ClusteringHP} will likely improve our performance further.

Secondly, our suggested retrieval at inference approach is fairly robust to alignment characteristics, and Tranception performs well with retrieval of just the nearest homologs. Our model can make use of deep alignments when they are available, and small, or no alignments when need be, resulting in both high-performance and broad scope. This is a significant advantage over alignment-based methods like EVE which require deep enough alignments to capture the complex relationships across residues in the protein sequence of interest.

Thirdly, unlike most existing mutation effect predictors, Tranception is able to handle insertions and deletions out-of-the-box, and outperforms prior baselines in that regime as well.

Fourthly, we find our approach to be robust across taxa and protein families, making it well suited to a broad range of tasks. This includes predicting the effect of mutations in viruses, a key component of forecasting outbreaks, and predicting disease causing variants in humans, of value to both diagnosis and preventative care.

Finally, our autoregressive model is naturally suited to sequence generation and hence has great potential for protein design.

\section*{Acknowledgements}

We thank Lood Van Niekerk and Aaron Kollasch for their help running the DeepSequence and Wavenet baselines on the ProteinGym benchmark, the broader OATML group and Marks Lab for helpful discussions when writing this manuscript, and the Google Cloud Platform team for research credits and TRC access when developing and training our models.
P.N. is supported by GSK and the UK Engineering and Physical Sciences Research Council (EPSRC ICASE award no. 18000077). 
M.D., J.F. and J.M.H. are supported by the Chan Zuckerberg Initiative CZI2018-191853. 
A.G. is a Clarendon Scholar and Open Philanthropy AI Fellow.
D.S.M. holds a Ben Barres Early Career Award by the Chan Zuckerberg Initiative as part of the Neurodegeneration Challenge Network, CZI2018-191853.
Y.G. holds a Turing AI Fellowship (Phase 1) at the Alan Turing Institute, which is supported by EPSRC grant reference V030302/1.

\bibliography{references}
\bibliographystyle{icml2022}

\newpage
\appendix
\onecolumn
\section*{Appendix}

\section{Glossary}
\label{glossary}

\textbf{Multiple Sequence Alignment (MSA): } a set of homologous protein sequences that are aligned in the position coordinate system of the seed sequence. They have been shown to capture information about the homology, phylogeny, and structure of the corresponding protein family \cite{Thompson1994CLUSTALWI, Thompson1997TheCW}. The depth of the MSA refers to the number of sequences it contains.

\textbf{Deep Mutational Scanning (DMS): } a technique which combines saturation mutagenesis with high-throughput functional screening to obtain a thorough description of the fitness landscape of the corresponding proteins \cite{fowler2014deep}. 

\textbf{Gain of scope: } increase in the number of distinct protein families that can be modeled. Proteins that have been historically difficult to model are the ones with shallow MSA or that are difficult to align (e.g., disordered proteins).

\section{Tranception model architecture and training details}

\subsection{Ablation studies}
\label{Appendix: Ablation studies}

Tranception is an autoregressive transformer architecture designed to explicitly promote head specialization and extraction of contiguous protein subsequence patterns, building on ideas introduced in Primer \cite{so2021primer} and Inception \cite{szegedy2014going}. We performed thorough ablations when developing Tranception and summarize the main variants tested in Table~\ref{Appendix - Table: Model architecture}. Our largest transformer model, Tranception L, has 700M parameters and is trained on UniRef100 \cite{10.1093/bioinformatics/btu739}. In early iterations we also experimented training our architecture on UniRef90 and UniRef50, clustered versions of UniRef100 at 90\% and 50\% similarity levels respectively, but observed superior performance from training on UniRef100 (see Appendix~\ref{Appendix:model_training} and Table~\ref{Table:Appendix_ablation_performance}).

\begin{table}[h]
    \centering
    \resizebox{\textwidth}{!}{
    \begin{tabular}{lcccccc}
        \toprule
        \textbf{Hyperparameter} &  \textbf{GPT2 S} & \textbf{Primer S} & \textbf{Tranception LS} & \textbf{Tranception S} & \textbf{Tranception M} & \textbf{Tranception L} \\
        \midrule
        Parameters & 85M & 85M & 85M & 85M & 300M & 700M \\
        \midrule
        Attention heads & 12 & 12 & 12 & 12 & 16 & 20\\
        Layers & 12 & 12 & 12 & 12 & 24 & 36\\
        Embedding size & 768 & 768 & 768 & 768 & 1,024 & 1,280 \\
        \midrule
        \multirow{2}{*}{\begin{tabular}{@{}l@{}}
Activation \\ function
\end{tabular}} & \multirow{2}{*}{\begin{tabular}{@{}l@{}}
GELU
\end{tabular}} & Squared & Squared & Squared & Squared & Squared\\
        & & ReLU  & ReLU & ReLU & ReLU & ReLU\\
        \midrule
        \multirow{2}{*}{\begin{tabular}{@{}l@{}}
Position \\ encoding
\end{tabular}} & Learned & Learned & Learned & Grouped & Grouped & Grouped \\
        & embedding & embedding  & embedding & ALiBi & ALiBi & ALiBi\\
        \bottomrule
    \end{tabular}
    }
    \caption{\textbf{Characteristics of different model variants used in ablations.} All models had a max context length of 1024 tokens, and we use the default dropout value of 0.1 in all variants. Tranception LS differs from Tranception S by the use of learned position embeddings instead of Grouped ALiBi.}
    \label{Appendix - Table: Model architecture}
\end{table}

To decide between different architecture options while not overfitting these decisions to our benchmark, we selected a small yet representative subset of DMS assays in the ProteinGym substitution benchmark (10 out of 87 substitution DMS assays):
\begin{multicols}{2}
\begin{itemize}[noitemsep,topsep=0pt]
    \item BLAT ECOLX \cite{jacquier_capturing_2013}
    \item CALM1 HUMAN \cite{weile_framework_2017}
    \item CCDB ECOLI \cite{tripathi_molecular_2016}
    \item DLG4 RAT \cite{mclaughlin_jr_spatial_2012}
    \item PA I34A1 \cite{wu_functional_2015}
    \item Q2N0S5 9HIV1 \cite{haddox_mapping_2018}
    \item RL401 YEAST  \cite{roscoe_analyses_2013}
    \item SPG1 STRSG  \cite{olson_comprehensive_2014}
    \item SPIKE SARS2 \cite{starr_deep_2020}
    \item TPOR HUMAN \cite{bridgford_novel_2020}
\end{itemize}
\end{multicols}
Together, these 10 assays cover the different taxa (3 viral proteins, 4 human and other eukaryote proteins, 3 prokaryote proteins), mutation depths (3 low, 4 medium and 3 high as per the classification described in Table\ref{Results - Table: Performance by MSA depth}) and include one assay with multiple mutants (which matches the overall proportion of assays with multiple mutants within ProteinGym).
Downstream performance of the different ablations on this validation set, and the overall substitution set are reported in Table~\ref{Table:Appendix_ablation_performance}.

\begin{table} 
\begin{center}
\begin{tabular}{ccccc}

\toprule

\multirow{2}{*}{\textbf{Model variant}} & \multirow{2}{*}{\textbf{Training data}} & \multirow{2}{*}{\textbf{Position encoding}} & \textbf{Spearman}  & \textbf{Spearman} \\
& & & \textbf{validation set} & \textbf{full set} \\
\midrule
GPT2 S & Uniref100 & Learned embedding & 0.324 & 0.320
\\
Primer S & Uniref100 & Learned embedding & 0.314 & 0.315
\\
Tranception LS & Uniref100 & Learned embedding & 0.330 & 0.333
\\
\midrule
Tranception S & Uniref100 & Grouped ALiBi & 0.344 & 0.335
\\
Tranception S & Uniref90 & Grouped ALiBi & 0.264 & 0.275
\\
Tranception S & Uniref50 & Grouped ALiBi & 0.248 & 0.247
\\
\midrule
Tranception M & Uniref100 & Grouped ALiBi & 0.358 & 0.376
\\
Tranception L & Uniref100 & Grouped ALiBi & \textbf{0.399} & \textbf{0.404}
\\
\bottomrule

\end{tabular}

\end{center}

\caption{\textbf{Performance of the different model variants in ablation studies.} Performance is measured via Spearman's rank correlation $\rho$ between model scores and experimental measurements, following the approach discussed in \ref{Appendix:performance_reporting}. Retrieval inference is excluded from this analysis. Model selection is performed on the validation set described in Appendix~\ref{Appendix: Ablation studies}.}
\label{Table:Appendix_ablation_performance}

\end{table}

\subsection{Data processing}
\label{Appendix: Data processing and augmentations}

Except for the two ablations focusing on UniRef50 and UniRef90, all models are trained on UniRef100. We perform very mild filtering steps of the data to remove fragments and low quality sequences, and preserve as much sequence diversity as possible. For each UniRef100 sequence cluster, we map the corresponding UniRef50 cluster which pools together sequences within 50\% similarity from one another \footnote{More precisely, UniRef100 is first clustered at the 90\% identity to generate UniRef90. Cluster representatives in UniRef90 are then clustered at 50\% identity to yield UniRef50.}. We use 99\% of the data ($\sim249$ million sequences) for training and set aside 1\% of the data for validation ($\sim2.5$ million sequences). All singletons at the UniRef50 cluster level are removed (eg., isolated fragments). We further exclude from the training and validation datasets all sequences that contained the infrequent Pyrrolysine (O) or Selenocysteine (U) amino acids, or with two or more consecutive indeterminate amino acids X to remove lower quality sequences. The remaining indeterminate amino acids (X, B, J, Z) are kept at train time and randomly imputed as follows: X is imputed to any of the 20 standard amino acids, B to either D (Aspartic acid) or N (Asparagine), J to either I (Isoleucine) or L (Leucine), Z to either E (Glutamic acid) or Q (Glutamine). All sequences with indeterminates are excluded from the validation set.
Table~\ref{Appendix - Table: Uniref100 statistics after filtering} recapitulates key statistics of sequences in UniRef100 after applying these filtering criteria. The observed distribution of sequences guided our choice for the maximum context length of 1,024 tokens for our transformer models, as it allows to handle $98\%$ of protein sequences in UniRef100 without truncation.

\begin{table}
\parbox{.45\linewidth}{
    \centering
    \begin{tabular}{cc}
        \toprule
        \textbf{Metric} &  \textbf{Value} \\
        \midrule
        Number of sequences & 249M \\
        \midrule
        Max sequence length & 40,921 \\
        $95^{th}$ percentile of length & 939 \\
        $75^{th}$ percentile of length & 470 \\
        Median sequence length & 314 \\
        $25^{th}$ percentile of length & 198 \\
        $5^{th}$ percentile of length & 92 \\
        Min sequence length & 12 \\
        \bottomrule
    \end{tabular}
    \caption{\textbf{High level statistics of protein sequences in UniRef100 after preprocessing.} About 98\% of sequences in UniRef100 have length lower than 1,024.}
    \label{Appendix - Table: Uniref100 statistics after filtering}
}
\hfill
\parbox{.45\linewidth}{
    \centering
    \begin{tabular}{cc}
        \toprule
        \textbf{Hyperparameter} &  \textbf{Value} \\
        \midrule
        Training steps & 150k \\
        Batch size & 1,024 \\
        Peak learning rate & $3 * 10^{-4}$ \\
        Weight decay & $10^{-4}$ \\
        Optimizer & AdamW \\
        \bottomrule
    \end{tabular}
    \caption{\textbf{Model training hyperparameters.}}
    \label{Appendix - Table: Model training}
}
\end{table}

\subsection{Model training}
\label{Appendix:model_training}

All model variants are trained for 150k steps, with a batch size of 1,024 sequences. During training, we reverse sequences at random and truncate sequences if longer than the maximum context size as per the scoring scheme details described in Appendix~\ref{Appendix:scoring_protein_sequences}.
We train with the AdamW optimizer \cite{Loshchilov2019DecoupledWD}, with a learning rate schedule annealed over the first 10k steps up to the maximum value ($3 * 10^{-4}$), and then linearly decreased until the end of training. Other training hyperparameters are summarized in Table~\ref{Appendix - Table: Model training}. In terms of computing resources, small architectures are trained on 8 V100 GPUs for $\sim$1 week, medium architectures with 32 V100 GPUs for $\sim$1 week, and our largest model, Tranception L, is trained on 64 A100 GPUs for $\sim$2 weeks. 

\begin{figure} [!tbp]
  \centering
  \begin{minipage}[b]{0.49\textwidth}
    \includegraphics[width=\textwidth]{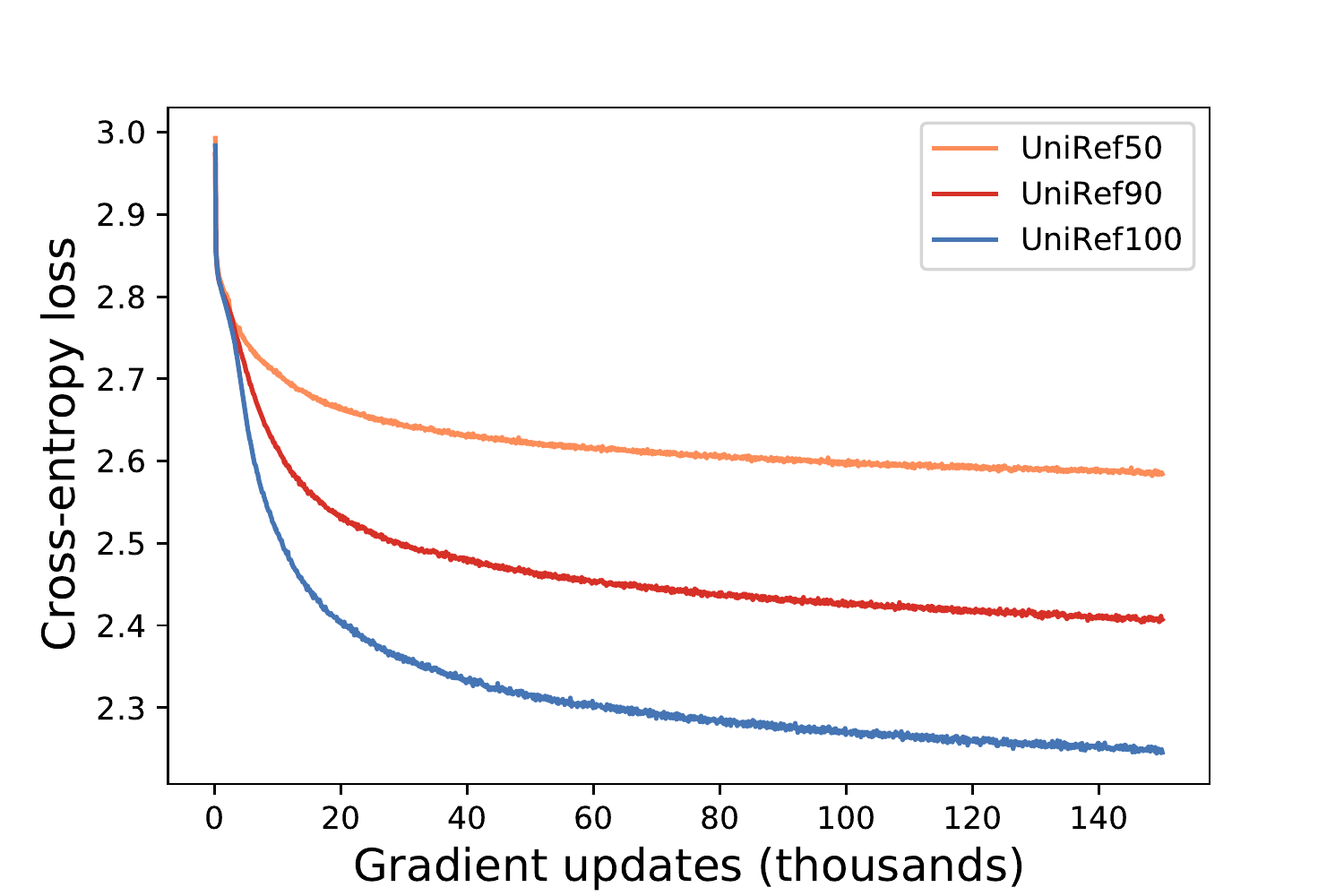}
  \end{minipage}
  \begin{minipage}[b]{0.49\textwidth}
    \includegraphics[width=\textwidth]{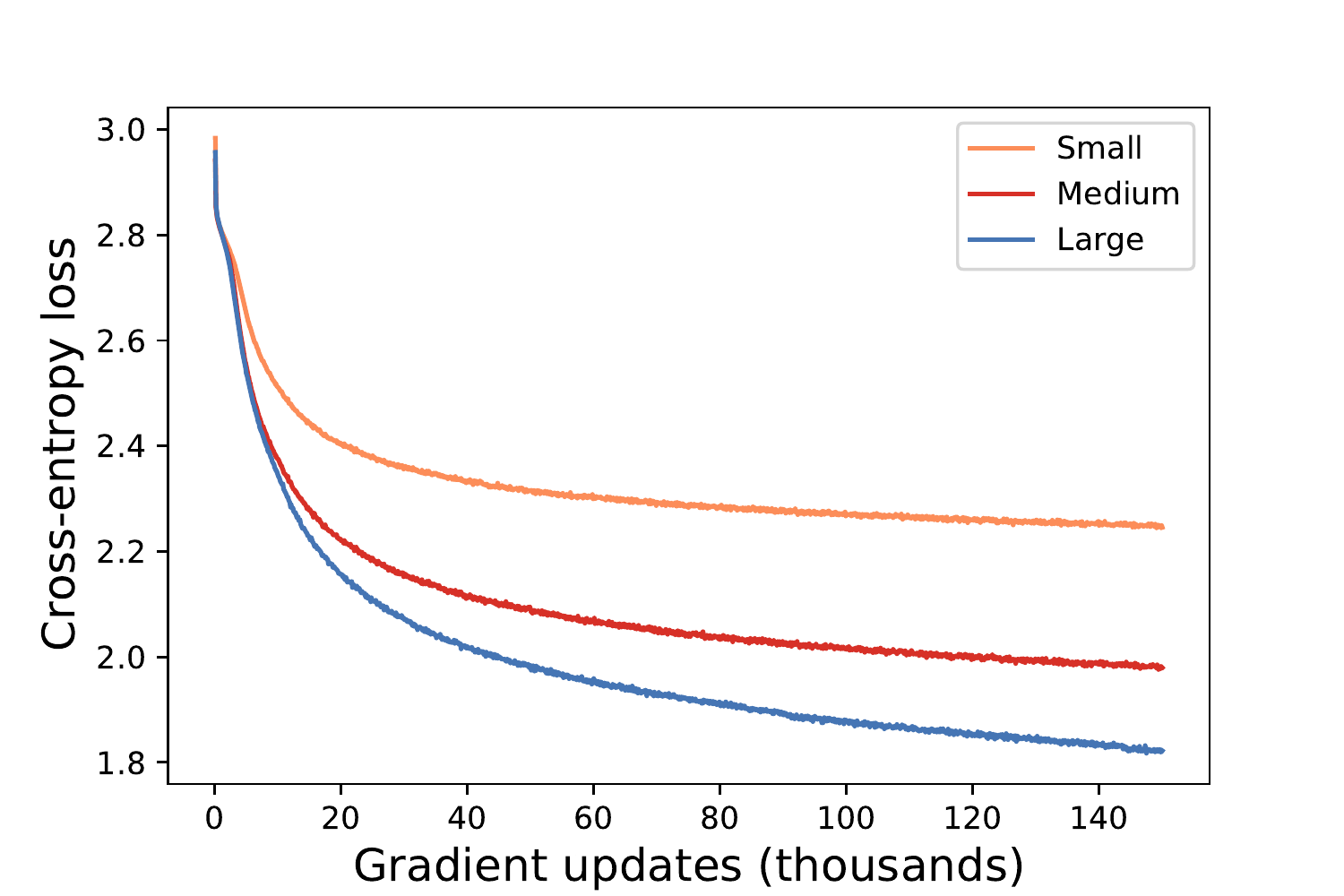}
  \end{minipage}
  \caption{\textbf{Cross-entropy loss Vs number of gradient steps} Left: Tranception S architecture trained on different UniRef datasets (UniRef50, UniRef90, UniRef100). Right: Small (S), Medium (M) or Large (L) Tranception architectures trained on UniRef100.}
  \label{Fig:Appendix_loss_curves}
\end{figure}

We provide the training loss curves of the different architecture variants in Fig.~\ref{Fig:Training_loss_GPT2_Primer_Tranception_M} and Fig.~\ref{Fig:Appendix_loss_curves}. While trained beyond the optimal number of steps for transformer models used in NLP tasks \cite{Kaplan2020ScalingLF}, our larger networks still appear to be undertrained. 
We note that cross-entropy is not necessarily a good indicator of downstream performance when comparing datasets with very different characteristics. For instance, UniRef50 is by design less redundant than UniRef100, thus we expect the loss to be typically lower on the latter (as can be seen in Fig.\ref{Fig:Appendix_loss_curves}). It so happens that, for the particular task we are interested in (fitness prediction), the more granular UniRef100 both leads to lower cross-entropy loss and higher downstream task performance (Table \ref{Table:Appendix_ablation_performance}), hence we trained our larger models on that dataset.

\subsection{Scoring protein sequences}
\label{Appendix:scoring_protein_sequences}

\paragraph{Mirrored sequences}
While protein sequences are not strictly invariant to mirroring (ie., the same sequence of amino acids assembled in the left to right order may not lead to the same 3D structure if assembled from right to left), prior autoregressive transformer architectures (eg., \cite{madani2020progen}) have proposed to augment their training dataset by including all sequences and their reverse.
In this work we further investigate the benefits of using each sequence and its reverse \emph{at inference} when predicting fitness (\S~\ref{Section: Tranception - Scoring sequences}). To remove potential discrepancies between training and inference, we apply a data augmentation similar to that of \citet{madani2020progen} at training time and randomly reverse a subset of sequences in each batch (each sequence in the batch is flipped with a probability 0.5).
We find that averaging the log-likelihood ratios obtained by scoring each sequence and its mirror image leads to superior downstream task performance, with or without retrieval (Table~\ref{Table:Appendix_mirrored_sequence_scoring}).

\begin{table}
    \centering
    \begin{tabular}{llccc}
        \toprule
        \textbf{DMS set} & \textbf{Model} & \textbf{Unidirectional scoring} & \textbf{Bidirectional scoring} \\
        \midrule
        \multirow{2}{*}{Validation set} & Tranception (w/o retrieval) & 0.376 & 0.399 \\
        & Tranception (w/ retrieval) & 0.447 & 0.452 \\
        \midrule
        \multirow{2}{*}{Full set} & Tranception (w/o retrieval) & 0.376 & 0.401 \\
        & Tranception (w/ retrieval) & 0.432 & 0.445 \\
        \bottomrule
    \end{tabular}
    \caption{\textbf{Comparison of unidirectional Vs bidirectional scoring} Performance is measured via Spearman's rank correlation $\rho$ between model scores and experimental measurements. Analysis is performed with Tranception L (with and without retrieval), and scores are reported on both the validation DMS set and the full DMS set. The bidirectional scoring is the average of log likelihood scores obtained by traversing the sequence in the canonical direction (left to right) and the reverse direction (right to left).}
    \label{Table:Appendix_mirrored_sequence_scoring}
\end{table}

\paragraph{Scoring window}
When scoring protein sequences that are longer than the maximum context length of the model (1,024 amino acids in our model), we have the choice to either truncate the sequence or combine predictions from slided (overlapping or non-overlapping) windows of the sequence. We observe very little performance difference between the different approaches in terms of average Spearman's rank correlation with validation DMS assays measurements, and therefore leveraged the former for simplicity. For single mutations, the optimal scoring window is selected so as to maximize the context available around that mutation for prediction when scoring the sequence from left to right and right to left. When scoring multiple mutants, we applied a very similar approach, maximizing context around the barycenter of the various mutant positions.

\subsection{Retrieval}
\label{Appendix:retrieval}
Augmenting the autoregressive predictions at each position via retrieval inference only occurs at the positions covered (even partially) by the MSA -- outside of theses positions, we fully rely on autoregressive predictions. We compute the pseudocounts at each position of the alignment via weighted Laplace smoothing \cite{Jurafsky2008SpeechAL}, with a small smoothing parameter ($10^{-5}$). As discussed in \S~\ref{Section: Retrieval}, sequence are weighted as per the procedure described in \cite{hopf2017mutation} and we fully ignore gaps in the MSA when computing the pseudocounts.

We optimize the retrieval inference weight $\alpha$ from equation~\ref{equation:log_proba_two_modes} via linearly-spaced grid search between 0.0 (no retrieval) and 1.0 (full retrieval on covered positions, autoregressive only otherwise). As per the results in table~\ref{Appendix - Table: Retrieval rate optimization}, the optimal $\alpha$ value on the validation DMS set is 0.6. Except for the analysis discussed in this section, whenever retrieval is used in this paper,it is with this 0.6 retrieval inference weight.

\label{Appendix: Retrieval rate optimization}

\begin{table}[H]
    \centering
    \begin{tabular}{cccccccccccc}
    \toprule
 \multirow{2}{*}{\textbf{DMS set}} & \multicolumn{11}{c}{\textbf{Retrieval inference weight $\alpha$}} 
\\
& \textbf{0.0} & \textbf{0.1} & \textbf{0.2} & \textbf{0.3} & \textbf{0.4} & \textbf{0.5} & \textbf{0.6} & \textbf{0.7} & \textbf{0.8} & \textbf{0.9} & \textbf{1.0}
\\
\midrule
Validation set & 0.399 & 0.412 & 0.424 & 0.435 & 0.444 & 0.450 & \textbf{0.452} & 0.449 & 0.443 & 0.431 & 0.397
\\
Full set & 0.401	& 0.410	& 0.419	& 0.428	& 0.435	& 0.441	& 0.445	& \textbf{0.446}	& 0.442	& 0.432	& 0.404	
\\
\bottomrule
    \end{tabular}
    \caption{\textbf{Retrieval inference weight optimization.} We perform a linearly-spaced grid search for $\alpha$ on our validation DMS set and obtain an optimal rate of 0.6.}
    \label{Appendix - Table: Retrieval rate optimization}
\end{table}

\section{Multiple Sequence Alignments}
\label{Appendix:MSA_creation}

For every assay in ProteinGym, we build MSAs of the corresponding protein by performing five search iterations of the profile HMM homology search tool Jackhmmer \cite{eddy2011accelerated} on the UniRef100 database of non-redundant proteins \cite{10.1093/bioinformatics/btu739}, downloaded on November 24 2021. We explore a range of 9 bit score thresholds, from 0.1 to 0.9. We subsequently select the alignment with the highest number of significant Evolutionary Couplings (ECs) \cite{hopf2014sequence}. To be consistent in our approach across all assays, we build a single MSA for each protein, and do not investigate domain-specific alignments where relevant. As noted in Appendix~\ref{Appendix:domain_specific_MSAs}, crafting domain-specific alignments, when these domains are known, may help further increase performance of the different models. 
Characteristics of the different MSAs (eg., number of sequences, selected bit score) we obtained with the above process are provided in the ProteinGym reference file available in our \href{https://github.com/OATML-Markslab/Tranception}{GitHub repository}.

\section{Baselines}
\label{Appendix:baselines}

Except for the enhancements we discuss in this section, we use the official codebases for all baselines included in \S~\ref{Section: Results}. For the Site independent and EVmutation models we use the \href{https://github.com/debbiemarkslab/EVcouplings}{EVcouplings library}. For Wavenet, we use the code made available in the \href{https://github.com/debbiemarkslab/SeqDesign}{SeqDesign} repository.
All alignment-based models are trained on the MSAs we obtain via the process discussed in Appendix~\ref{Appendix:MSA_creation}, and the same MSAs are used at inference time with MSA transformer or for retrieval in Tranception.

\subsection{ESM-1v and MSA Transformer}

We start from the \href{https://github.com/facebookresearch/esm}{official ESM codebase}, and extend it as follows to support:
\begin{itemize}[noitemsep,topsep=0pt]
    \item the scoring of multiple mutants -- as discussed in \citet{Meier2021.07.09.450648} this is achieved by independently summing the effects of each single mutations that comprise the multiple mutant;
    \item the scoring of sequences that are longer than the context size of the ESM-1v and MSA Transformer models. We leverage the same routine we developed for Tranception to select the optimal scoring window (see Appendix~\ref{Appendix:scoring_protein_sequences});
    \item the weighted sampling of sequences of an input MSA in the MSA Transformer. We compute sequence weights with the same procedure used for retrieval inference in Tranception, ie. based on the procedure described in \citet{hopf2017mutation}.
\end{itemize}

When scoring with ESM-1v or MSA Transformer, we use the masked-marginals approach introduced in \citet{Meier2021.07.09.450648}. This scoring heuristics has two main limitations: 1) as discussed above, scores for multiple mutants are computed as the sum of the effects from the individual single mutants comprising the multiple mutation, which leads to ignoring potentially important epistatic effects 2) masked scoring implies a fixed frame of reference in which the mutated position exists in the original wild-type sequence. Consequently these two models are unable to score indels with the masked-marginals heuristic.

As discussed in \S~\ref{Section: Results}, we compute scores reported in the main results tables are obtained with a single model seed for fair comparison across models, but we also report ensemble performance in Appendix~\ref{Appendix: Ensembling}. We use the most recent checkpoints made publicly available for each model (resp. esm1v\_t33\_650M\_UR90S\_1 and esm\_msa1b\_t12\_100M\_UR50S at the time of writing).
For MSA transformer, we follow \cite{Meier2021.07.09.450648} and first filter sequences in the input MSA with HHFilter \cite{Steinegger2019HHsuite3FF} to ensure minimum coverage of 75\% and maximum sequence identity of 90\%, and then sample 384 sequences (with the weights discussed above).

\subsection{EVE and DeepSequence}
For both DeepSequence and EVE, we use the Pytorch implementation available in the official \href{https://github.com/OATML-Markslab/EVE}{EVE github repository} (we use the optimal parameters for EVE as per \citet{Frazer2021DiseaseVP}, and as per \citet{riesselman2018deep} for DeepSequence).

EVE and DeepSequence models are trained on protein-specific MSAs. Sequences with more than 50\% of gaps in the MSA are removed from training. More importantly, positions in the MSA with more than 30\% of gaps in the MSA are removed, such that EVE and DeepSequence models do not provide scores for mutations impacting these particular positions. When the full protein is difficult to cover, this sometimes results in a full domain of the protein being dropped (see the BRCA1 protein example in Appendix~\ref{Appendix:domain_specific_MSAs}). For that reason, when reporting performance results in \S~\ref{Section: Results} we compare all models on the exact same set of mutants, excluding the ones that are not scored by EVE and DeepSequence.
We nonetheless provide a full performance comparison on the entire set of mutants in the files available in our \href{https://github.com/OATML-Markslab/Tranception}{GitHub repository} for the models that provide scores for all mutants (see \S~\ref{Appendix:performance_analysis}).

Except for Wavenet, all alignment-based models are unable to score indels since they rely on the fixed coordinate system from the original MSA they have been trained on.

\section{Detailed performance analysis}
\label{Appendix:performance_analysis}

\subsection{Performance reporting methodology}
\label{Appendix:performance_reporting}

We report performance based on the 3 metrics described in \S~\ref{Section: Results}:
\begin{itemize}[noitemsep,topsep=0pt]
    \item \textbf{Spearman's rank correlation $\rho$ between model scores and DMS experimental measurements.} Since certain DMS assays are relatively difficult to model resulting in very low (and sometimes negative) $\rho$ values, we report the signed $\rho$ value instead of the absolute $\rho$ values reported in prior works \cite{riesselman2018deep,Meier2021.07.09.450648}. To that end, we adjust the signs of measured phenotypes where needed, to ensure consistency in the directionality across assays. In ProteinGym, a higher DMS score is always associated with higher fitness;
    \item \textbf{AUC between model scores and DMS experimental measurements.} This metric is particularly relevant when focusing on assays with a bimodal distribution of the measured phenotype. We binarize DMS measurements by setting the threshold manually when the assay is clearly bimodal and there is no ambiguity about the correct threshold value to select. We use a median cutoff in all other instances. We report the numerical value of the cutoff and the binarization method (median or manual) in the ProteinGym reference file (available in our \href{https://github.com/OATML-Markslab/Tranception}{GitHub repository});
    \item \textbf{Matthew's correlation coefficient (MCC) between model scores and DMS experimental measurements.} This complements the analysis obtained with AUC. DMS measurements are binarized with the same thresholds as for AUC. Predictions are binarized by using their median value as threshold.
\end{itemize}

In order to standardize measured outcomes as much as possible across assays, we further preprocess the raw DMS measurements as follows:
\begin{itemize}[noitemsep,topsep=0pt]
    \item \textbf{Silent mutations}: certain assays include nucleotide substitutions resulting in silent mutations. We remove these from our benchmark;
    \item \textbf{Duplicate mutations}: certain assays include duplicate mutants -- either nucleotide substitutions resulting in mutant repeats in the experiments, or indels resulting in identical protein sequences. We remove duplicates by averaging all DMS measurements across duplicate mutants;
    \item \textbf{Missing measurements}: mutants with missing assay measurement are dropped.
\end{itemize}

Finally, ProteinGym contains several DMS assays for the same protein, for a handful of proteins (eg., we include 4 assays for BLAT ECOLX, 4 assays for P53). While all different and important in their own right, these different assays have experimental measurements that are strongly correlated. Consequently, models that tend to do well for one of these assays relative to others, tend to perform well across all of the assays for the same protein. In order to remove the potential biases resulting from these correlated assays, we first aggregate all performance metrics at the Uniprot ID level and then compute the different average results in \S~\ref{Section: Results} (ie., when there 4 DMS assays for the same protein, each of these assays carries a weight of 0.25 in the aggregate results).

\subsection{Detailed results}
\label{Appendix:detailed_results}

\begin{figure}
    \centering
    \includegraphics{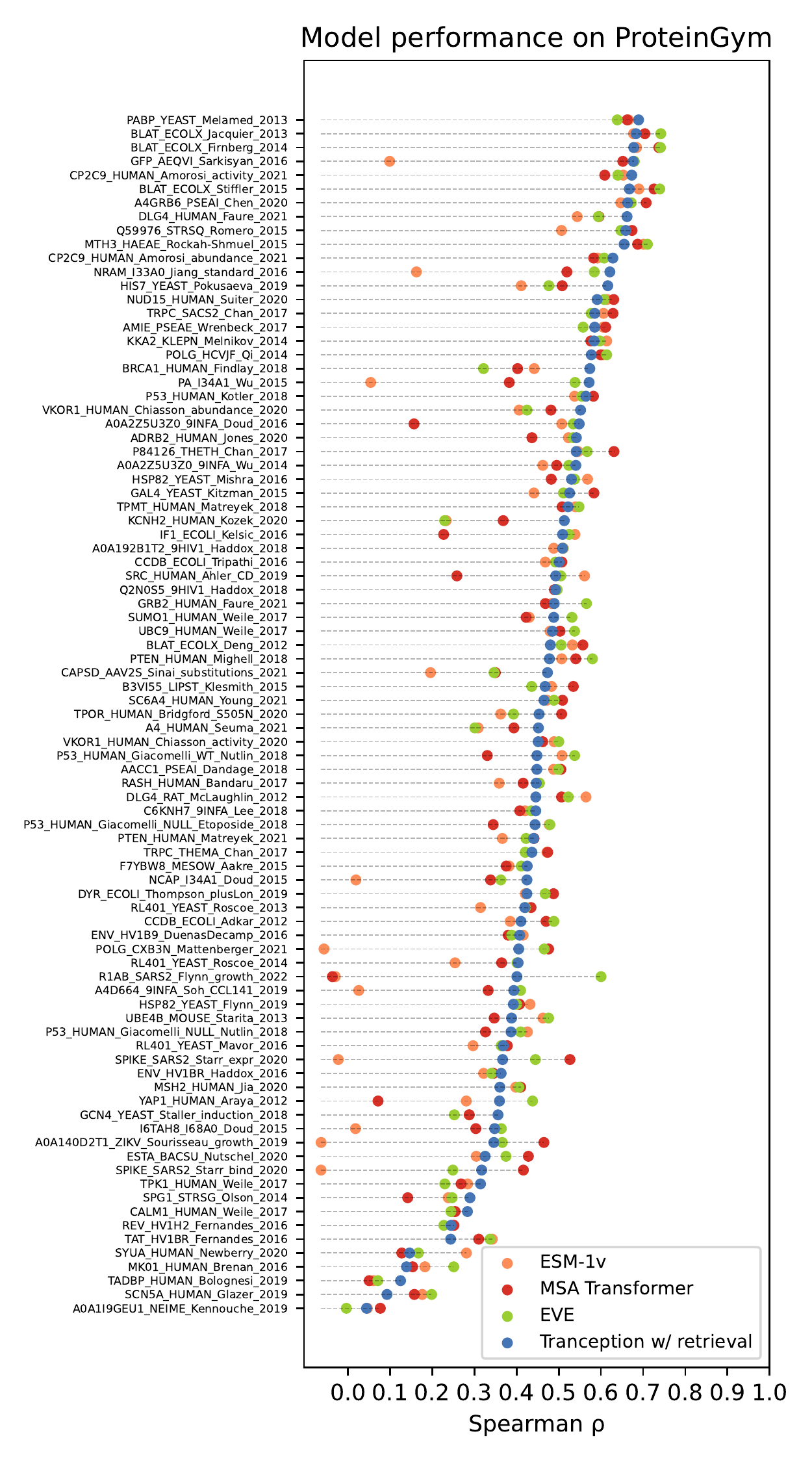}
    \caption{\textbf{Model performance on the ProteinGym substitution benchmark.} We report the DMS-level performance of Tranception with retrieval, ESM-1v \cite{Meier2021.07.09.450648}, MSA Transformer \cite{Rao2021.02.12.430858} and EVE \cite{Frazer2021DiseaseVP} for each DMS in the ProteinGym substitution benchmark. Performance is measured by the Spearman's rank correlation $\rho$ between model scores and experimental measurements.}
    \label{Fig. all_the_spearmans}
    
\end{figure}

We report below aggregated performance results by taxon for the different metrics described in Appendix~\ref{Appendix:performance_reporting} on the ProteinGym substitution benchmark: Spearman's rank in Table~\ref{Appendix_table:spearman_taxa_table}, AUC in Table~\ref{Appendix_table:AUC_taxa_table} and MCC in Table~\ref{Appendix_table:MCC_taxa_table}. Overall model ranking is the same across all performance metrics. 
The Spearman's rank correlation coefficients between model scores and DMS experimental measurements for Tranception, EVE, ESM-1v and MSA Transformer for each DMS assay in the substitution benchmark are shown in Fig.~\ref{Fig. all_the_spearmans}. Performance for the same metric for Tranception and Wavenet on the ProteinGym indel benchmark is provided in Table~\ref{Appendix_table:ProteinGym_indel_benchmark}.

Detailed performance tables for all models at the Uniprot ID level and at the DMS level for both the Protein substitutions and indels benchmarks are made available in the \href{https://github.com/OATML-Markslab/Tranception}{Tranception GitHub repository}.

\begin {table*}[t]

\begin{center}

\begin{tabular*}{\textwidth}{ @{\extracolsep{\fill}} llccccM{2.5cm}}

\toprule
\textbf{Model}  & \textbf{Model} & \multicolumn{5}{c}{\textbf{Spearman correlation by taxa category $\uparrow$}}  \\
\textbf{type} & \textbf{name} &  \textbf{Human} & \textbf{Other Eukaryote} & \textbf{Prokaryote} & \textbf{Virus} & \textbf{All} \\
\toprule
\multirow{5}{*}{\begin{tabular}[c]{@{}l@{}}
Alignment-  \\ based \\ models
\end{tabular}}
 & Site indep & 0.398 & 0.446 &	0.350 &	0.410 & 0.397 \\
 & Wavenet & 0.388 &	0.453 &	0.480 &	0.308 & 0.398 \\
 & Deepsequence & 0.391 &	0.482 &	0.487 &	0.350 & 0.415 \\
 & EVmutation & 0.405 &	0.475 &	0.484 &	0.380 & 0.427 \\
& EVE & 0.411 &	0.485 &	\textbf{0.497} &	\textbf{0.435} & 0.448 \\
\midrule
\multirow{4}{*}{\begin{tabular}{@{}l@{}}
Protein  \\ language \\ models
\end{tabular}}
 & ESM-1v & 0.394 &	0.420 &	0.482 &	0.216 & 0.371 \\
& MSA Transformer & 0.379 &	0.491 &	0.494 &	0.380 & 0.422 \\
& Tranception (w/o retrieval) & 0.369 &	0.441 &	0.453 &	0.396 & 0.406 \\
 & Tranception (w/ retrieval) & \textbf{0.426} &	\textbf{0.502} &	0.485 &	0.429 & \textbf{0.451} \\
\bottomrule
\end{tabular*}

\end{center}
\caption{\textbf{Average Spearman's rank correlation $\rho$ between model scores and experimental measurements by taxon.}}
\label{Appendix_table:spearman_taxa_table}
\end{table*}
\begin{table}[t]
\begin{center}
    \begin{tabular*}{\textwidth}{ @{\extracolsep{\fill}} llccccM{2.5cm}}
    \toprule
\textbf{Model}  & \textbf{Model} & \multicolumn{5}{c}{\textbf{AUC by taxa category $\uparrow$}}  \\
\textbf{type} & \textbf{name} &  \textbf{Human} & \textbf{Other Eukaryote} & \textbf{Prokaryote} & \textbf{Virus} & \textbf{All} \\
\toprule
\multirow{5}{*}{\begin{tabular}[c]{@{}l@{}}
Alignment-  \\ based \\ models
\end{tabular}}
 & Site indep & 0.732 & 0.768 &	0.696 &	0.720 & 0.725 \\
& Wavenet & 0.730 &	0.766 &	0.763 &	0.665 & 0.725 \\
& Deepsequence & 0.729 &	0.779 &	0.766 &	0.685 & 0.733 \\
 & EVmutation & 0.735 &	0.775 &	0.760 &	0.702 & 0.738 \\
& EVE & 0.742 &	0.782 &	\textbf{0.769} &	\textbf{0.732} & 0.751 \\
\midrule
\multirow{4}{*}{\begin{tabular}[c]{@{}l@{}}
Protein  \\ language \\ models
\end{tabular}}
 & ESM-1v & 0.734 &	0.749 &	0.762 &	0.620 & 0.713 \\
& MSA Transformer & 0.723 &	0.784 &	0.768 &	0.702 & 0.737 \\
& Tranception (w/o retrieval) & 0.716 &	0.762 &	0.745 &	0.709 & 0.728 \\
 & Tranception (w/ retrieval) & \textbf{0.749} &	\textbf{0.795} &	0.766 &	0.727 & \textbf{0.754} \\
\bottomrule
    \end{tabular*}
    
    \end{center}
    \caption{\textbf{Average AUC between model scores and experimental measurements by taxon.}}
    \label{Appendix_table:AUC_taxa_table}
\end{table}
\begin{table}
    \centering
    \begin{tabular*}{\textwidth}{@{\extracolsep{\fill}}llccccM{2.5cm}}
    \toprule
\textbf{Model}  & \textbf{Model} & \multicolumn{5}{c}{\textbf{MCC by taxa category $\uparrow$}}  \\
\textbf{type} & \textbf{name} &  \textbf{Human} & \textbf{Other Eukaryote} & \textbf{Prokaryote} & \textbf{Virus} & \textbf{All} \\
\toprule
\multirow{5}{*}{\begin{tabular}[c]{@{}l@{}}
Alignment-  \\ based \\ models
\end{tabular}} & Site indep & 0.323 & 0.344 &	0.279 &	0.311 & 0.312 \\
& Wavenet & 0.322 &	0.351 &	0.369 &	0.233 & 0.314 \\
& Deepsequence & 0.327 &	0.369 &	0.386 &	0.261 & 0.330 \\
 & EVmutation & 0.333 &	0.364 &	0.370 &	0.288 & 0.334 \\
& EVE & 0.338 &	0.366 &	0.387 &	\textbf{0.333} & 0.352 \\
\midrule
\multirow{4}{*}{\begin{tabular}[c]{@{}l@{}}
Protein  \\ language \\ models
\end{tabular}}
 & ESM-1v & 0.321 &	0.325 &	0.377 &	0.174 & 0.296 \\
& MSA Transformer & 0.311 &	0.388 &	\textbf{0.389} &	0.290 & 0.334 \\
& Tranception (w/o retrieval) & 0.302 &	0.349 &	0.350 &	0.298 & 0.319 \\
 & Tranception (w/ retrieval) & \textbf{0.348} &	\textbf{0.390} &	0.379 &	0.330 & \textbf{0.356} \\
\bottomrule
    
    \end{tabular*}
    \caption{\textbf{Average Matthew's correlation coefficient (MCC) between model scores and experimental measurements by taxon.}}
    \label{Appendix_table:MCC_taxa_table}
\end{table}

\begin {table*}[t]

\begin{center}
\begin{tabular}{lccc}

\toprule

\textbf{DMS assay} & \textbf{Wavenet} & \textbf{Tranception w/o retrieval} & \textbf{Tranception w/ retrieval} \\
\midrule

A0A1J4YT16 9PROT \cite{davidi2020highly} & 0.117 & 0.178	& \textbf{0.191}	\\
B1LPA6 ECOSM \cite{russ2020evolution}	& 0.385 & 0.321	& \textbf{0.415}	 \\
BLAT ECOLX \cite{gonzalez2019fitness} & \textbf{0.546}	& 0.296	& 0.357	 \\
PTEN HUMAN \cite{mighell_saturation_2018} & \textbf{0.699} &	0.563 &	0.598	 \\
CAPSD AAV2S \cite{sinai2021generative} & 0.457	& 0.549	& \textbf{0.586} \\
HIS7 YEAST \cite{pokusaeva_experimental_2019} & 0.680 &	\textbf{0.707} &	0.692	 \\
P53 HUMAN \cite{kotler_systematic_2018}	& 0.001 & 0.395	& \textbf{0.401} \\
\midrule
\textbf{Average}	& 0.412 & 0.430	& \textbf{0.463} \\

\bottomrule

\end{tabular}

\end{center}

\caption{\textbf{Average Spearman's rank correlation $\rho$ between model scores and experimental measurements on the ProteinGym indel benchmark.}}
\label{Appendix_table:ProteinGym_indel_benchmark}
\end{table*}

\subsection{MSA Filtering analyses}
\label{Appendix:filtering_analysis}

\begin{figure}[ht]
    \center
    \includegraphics[ width = 0.3 \linewidth, keepaspectratio]{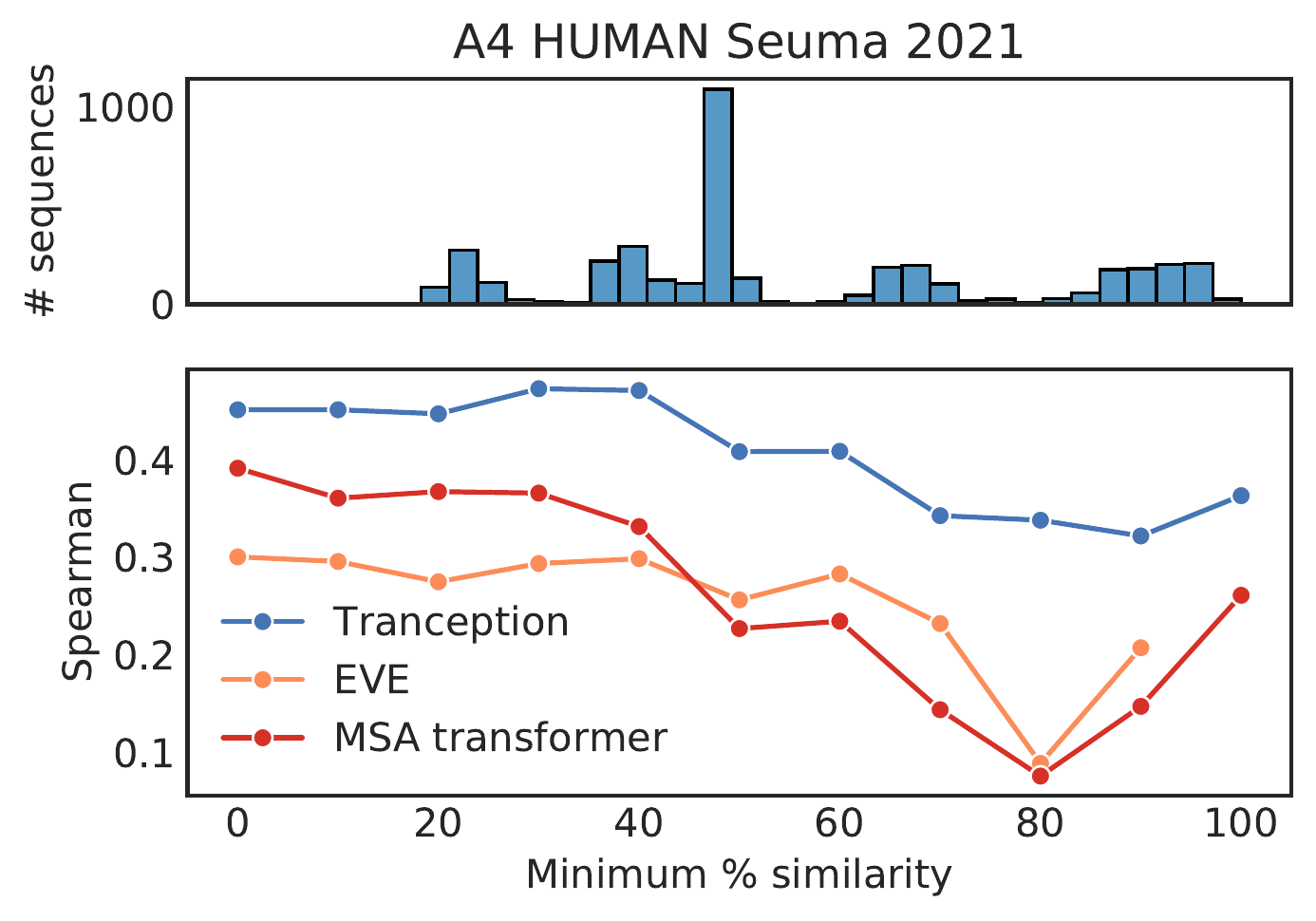}
    \hspace{0.01\linewidth}
    \includegraphics[ width = 0.3 \linewidth, keepaspectratio]{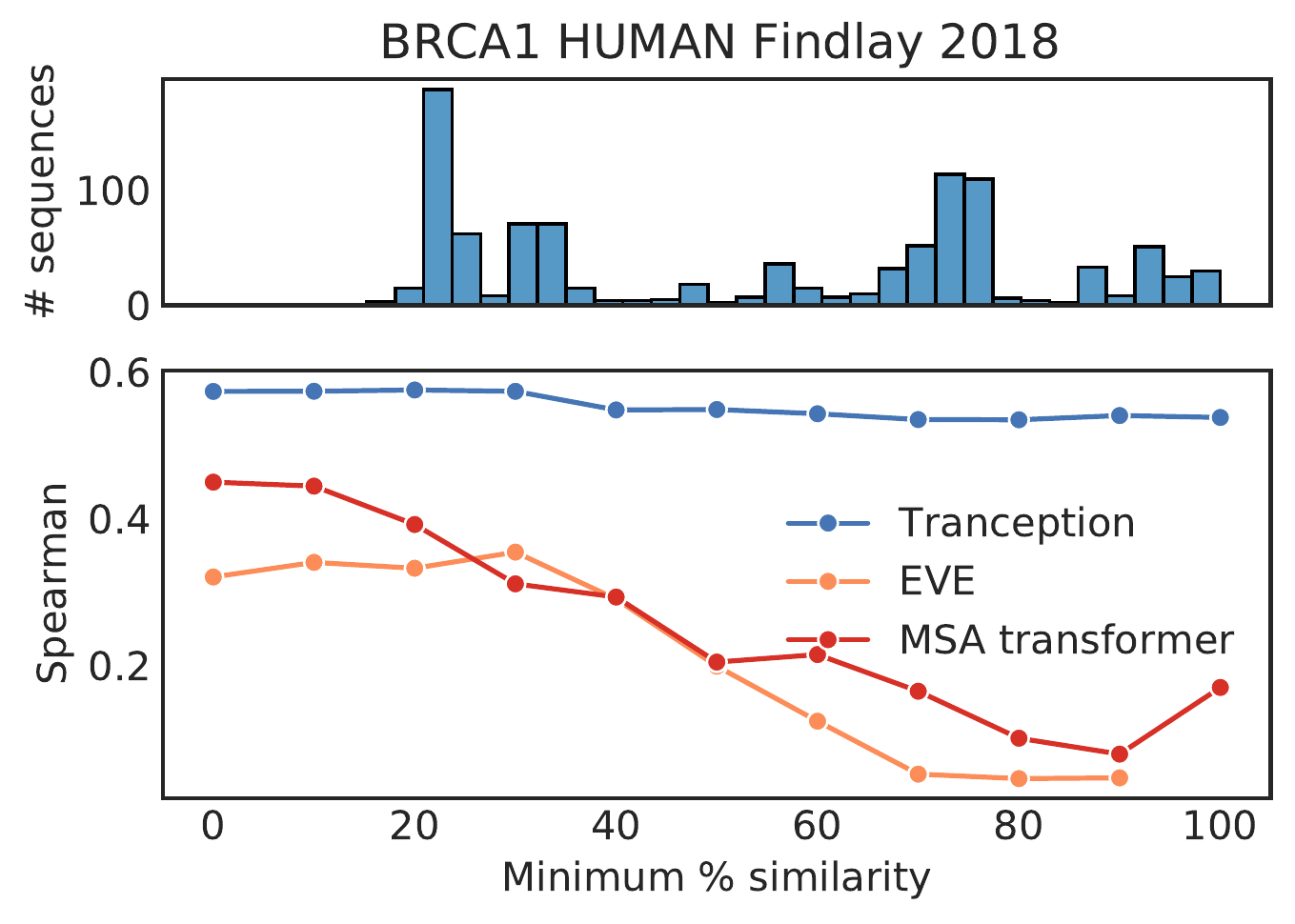}
    \hspace{0.01\linewidth}
    \includegraphics[ width = 0.3 \linewidth, keepaspectratio]{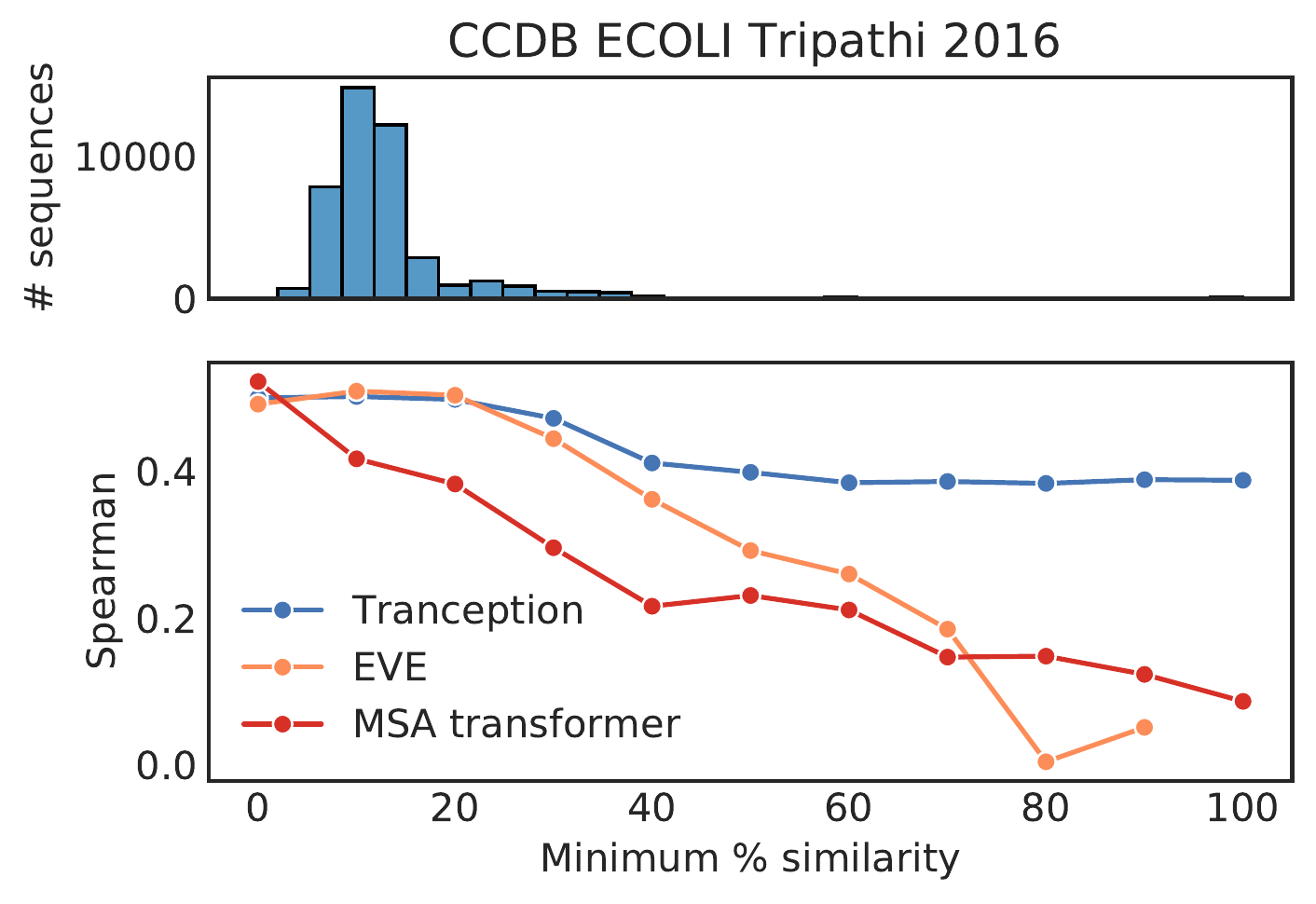}
    
    \vspace{0.03\textheight}
    
    \includegraphics[ width = 0.3 \linewidth, keepaspectratio]{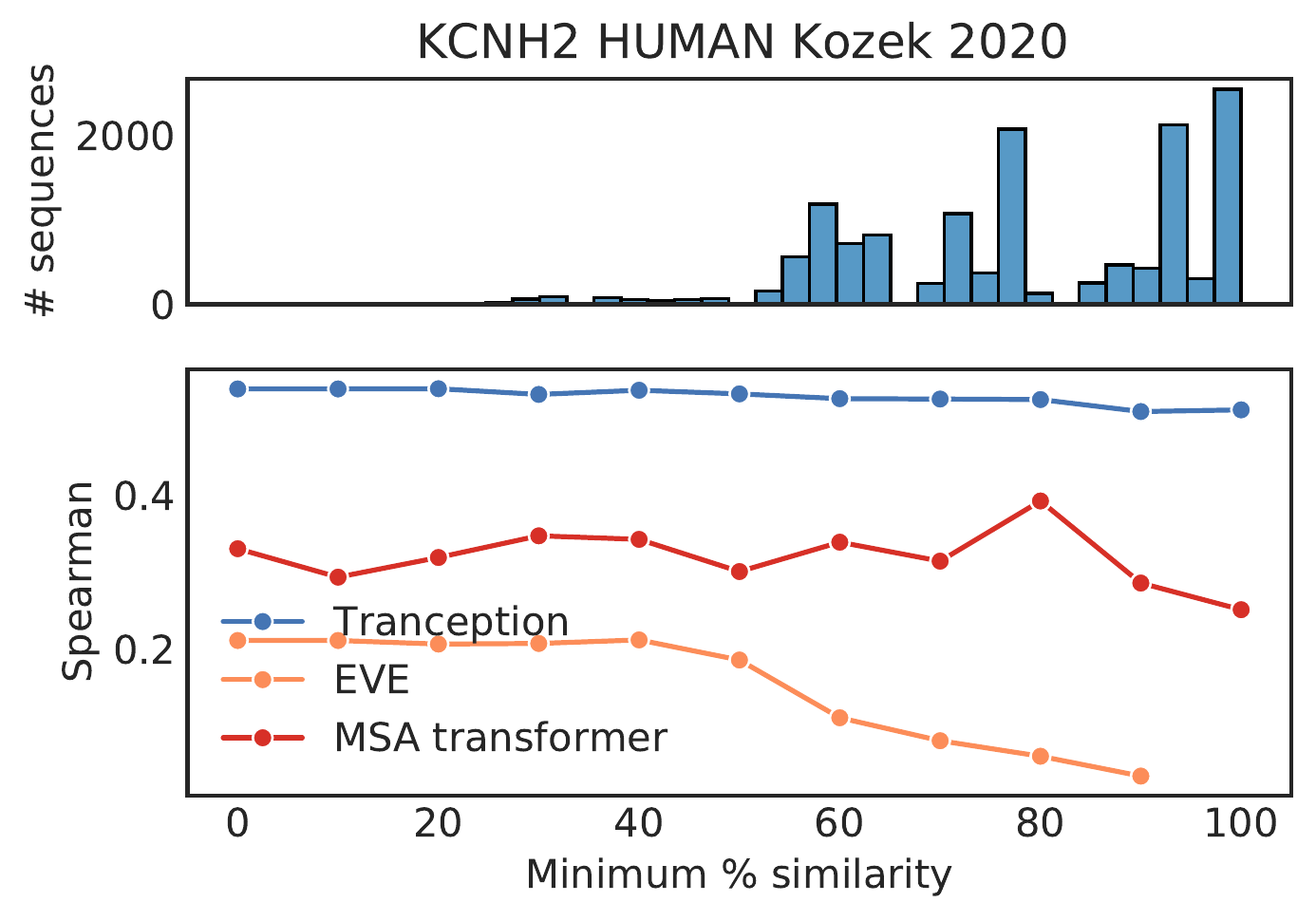}
    \hspace{0.01\linewidth}
    \includegraphics[ width = 0.3 \linewidth, keepaspectratio]{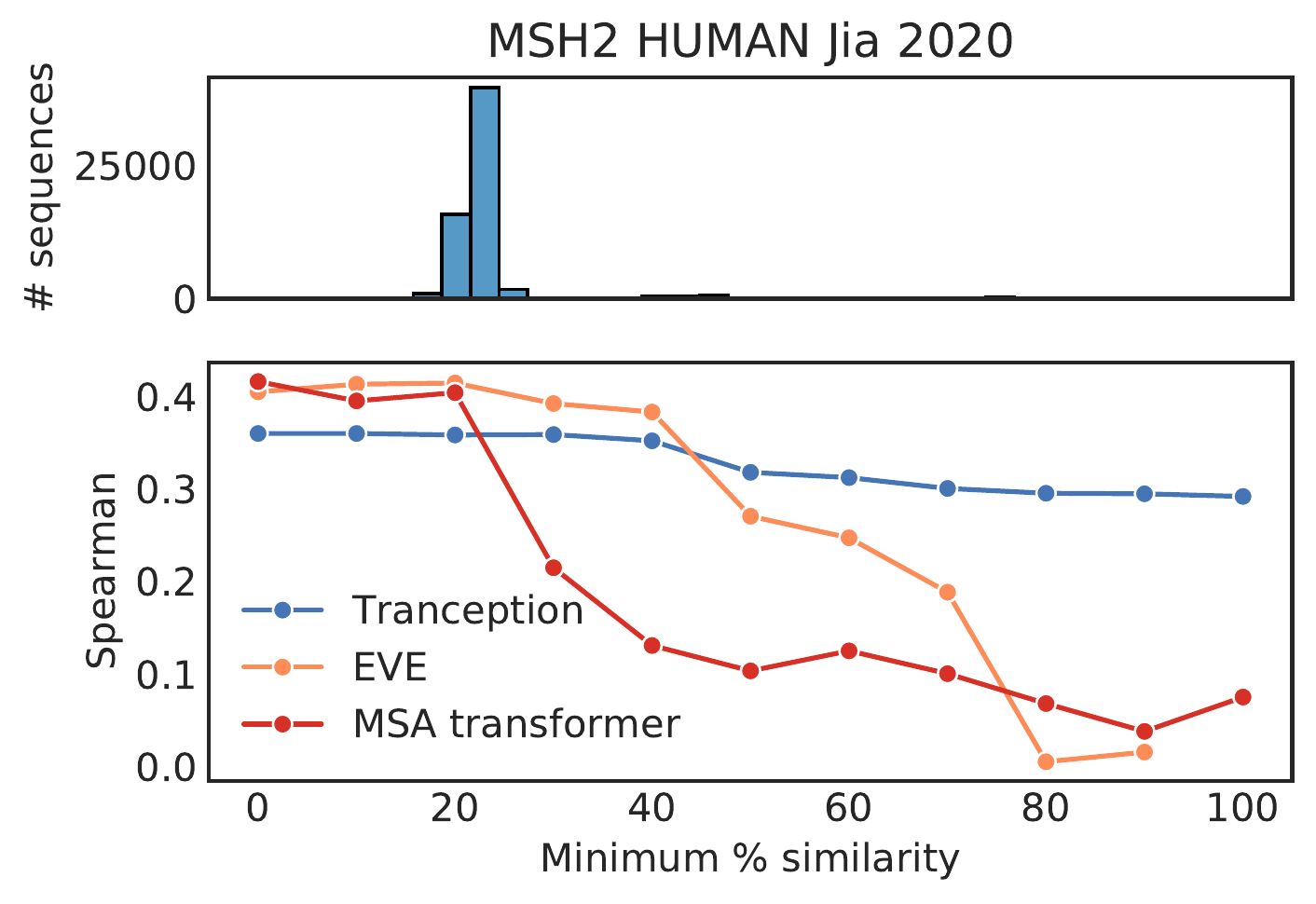}
    \hspace{0.01\linewidth}
    \includegraphics[ width = 0.3 \linewidth, keepaspectratio]{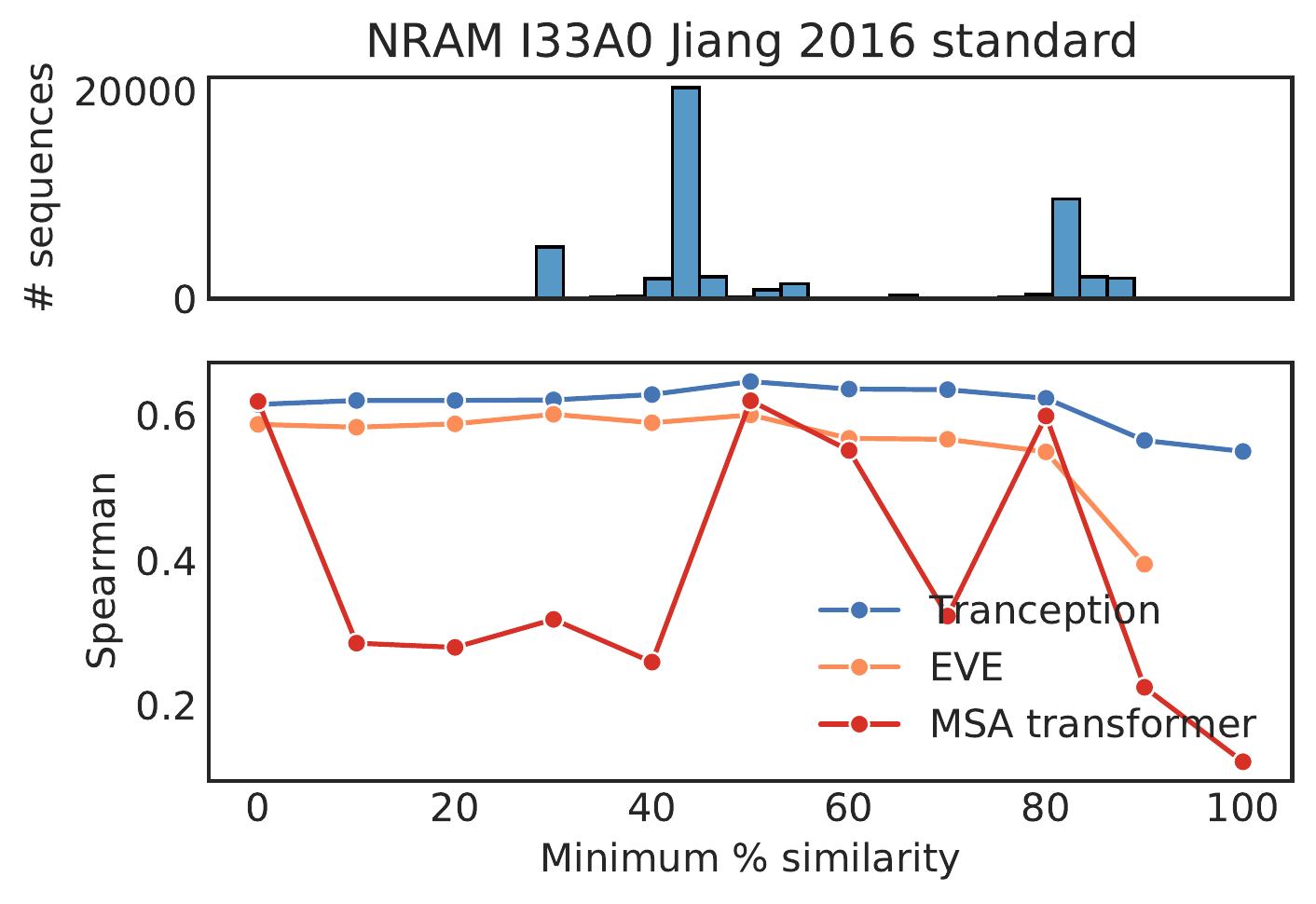}
    
    \vspace{0.03\textheight}
    
    \includegraphics[ width = 0.3 \linewidth, keepaspectratio]{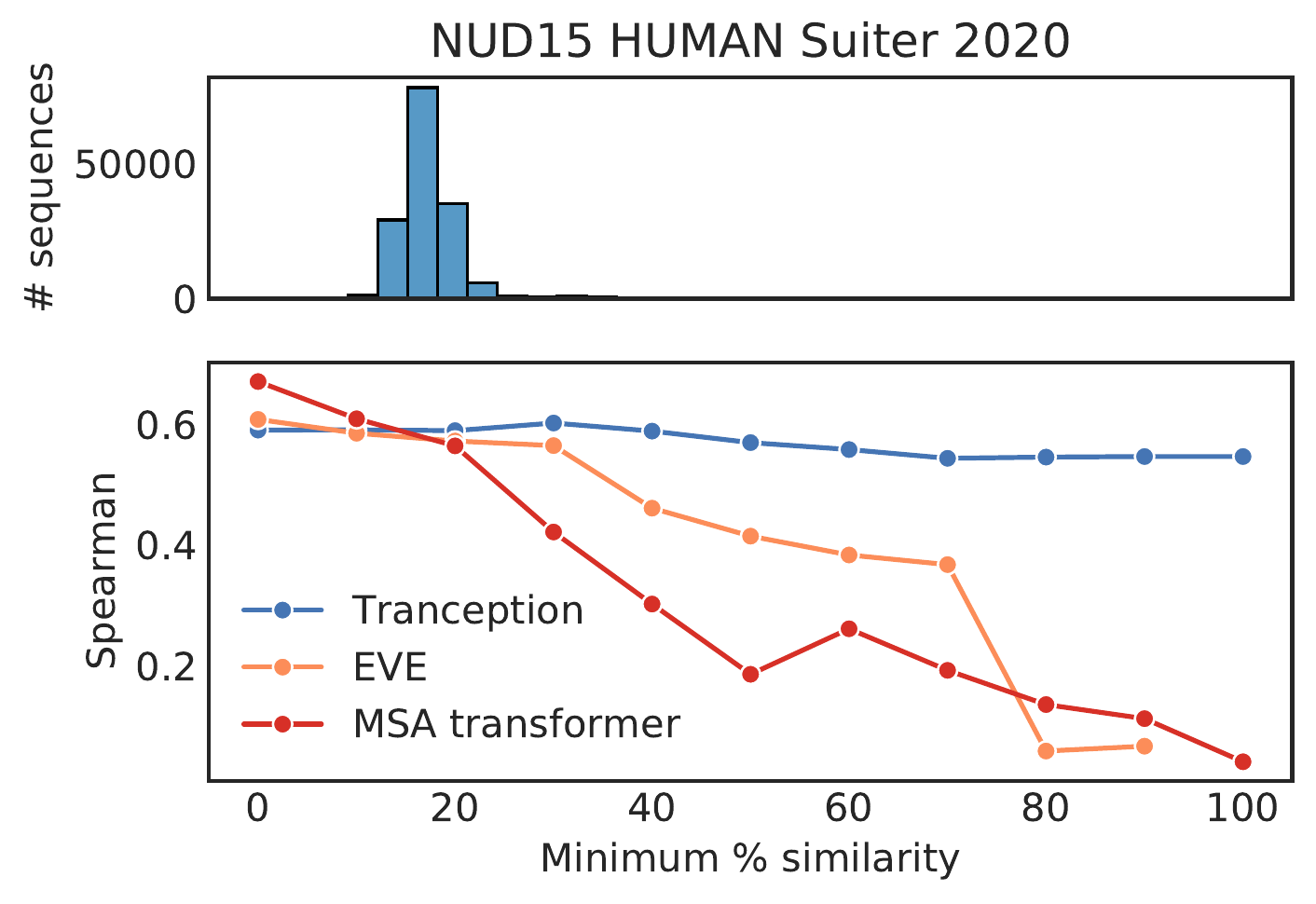}
    \hspace{0.01\linewidth}
    \includegraphics[ width = 0.3 \linewidth, keepaspectratio]{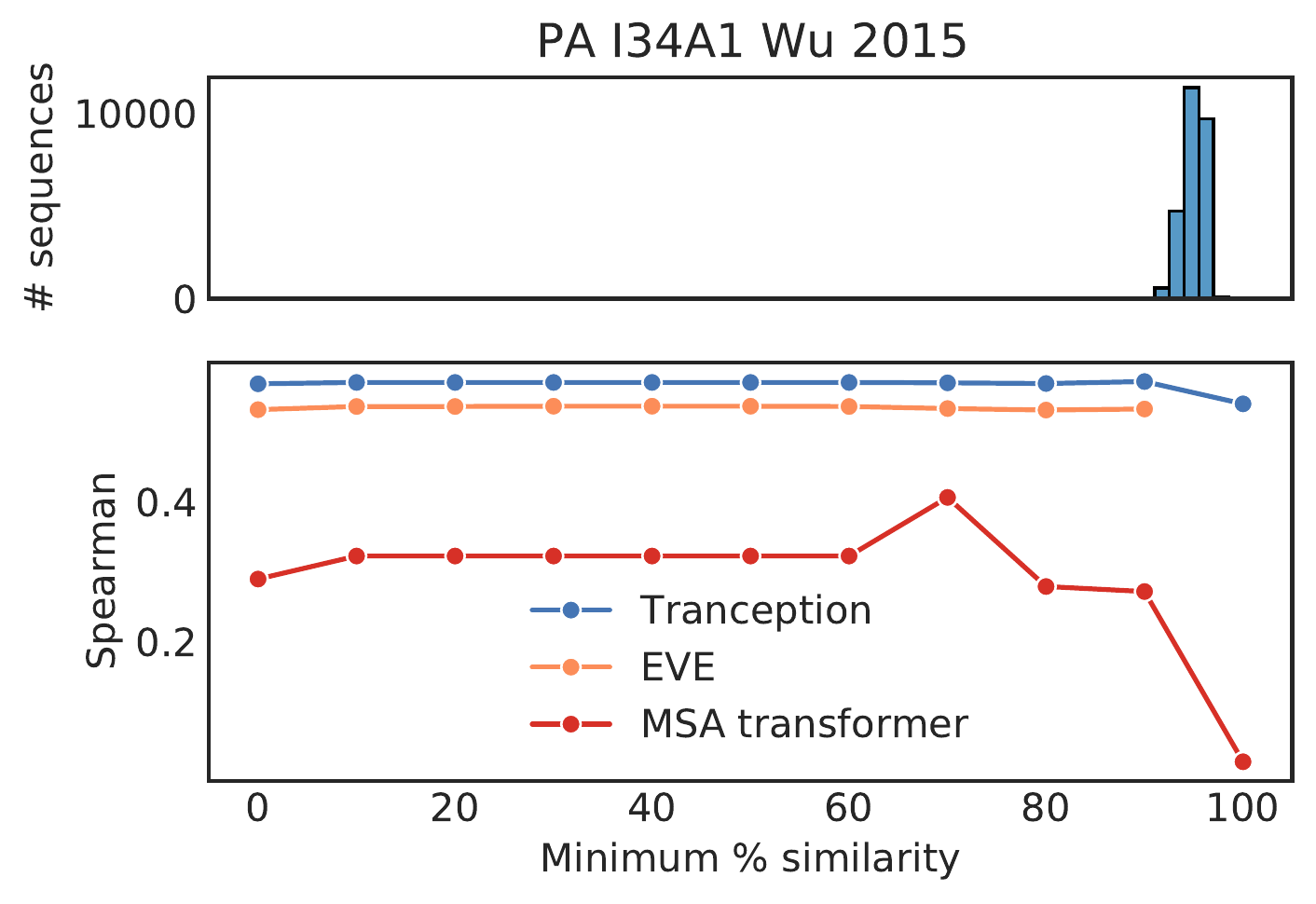}
    \hspace{0.01\linewidth}
    \includegraphics[ width = 0.3 \linewidth, keepaspectratio]{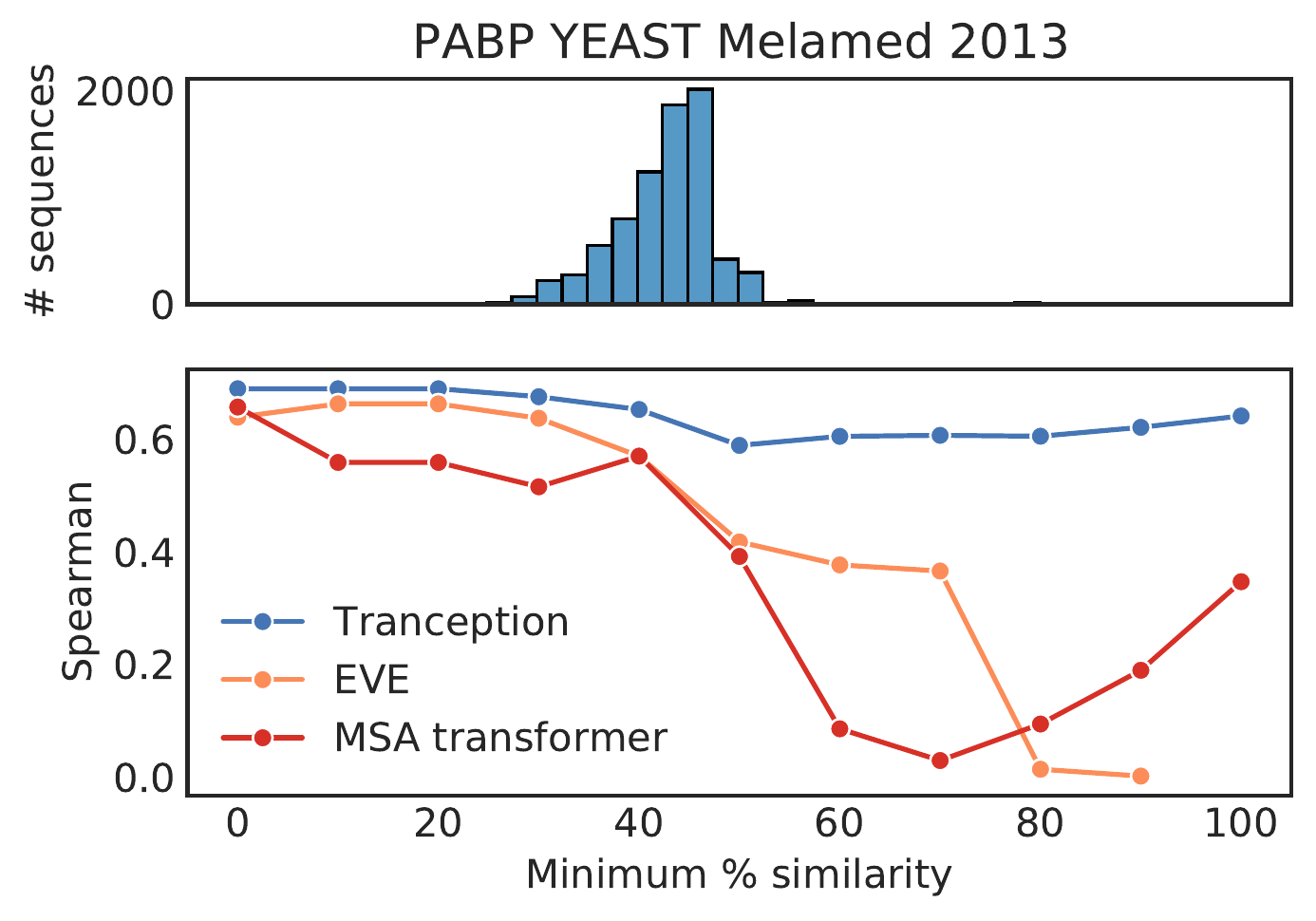}
    
    \vspace{0.03\textheight}
    
    \includegraphics[ width = 0.3 \linewidth, keepaspectratio]{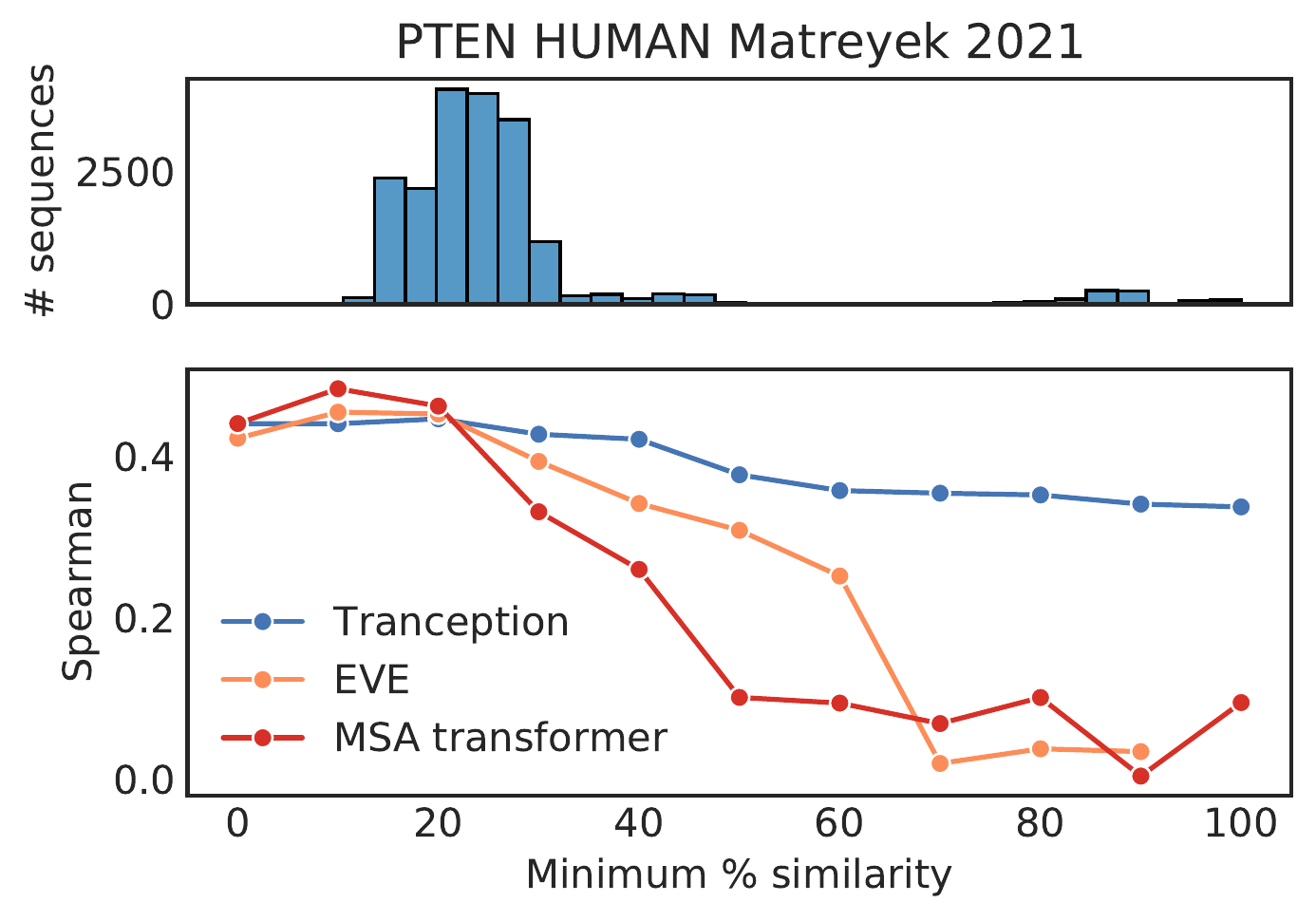}
    \hspace{0.01\linewidth}
    \includegraphics[ width = 0.3 \linewidth, keepaspectratio]{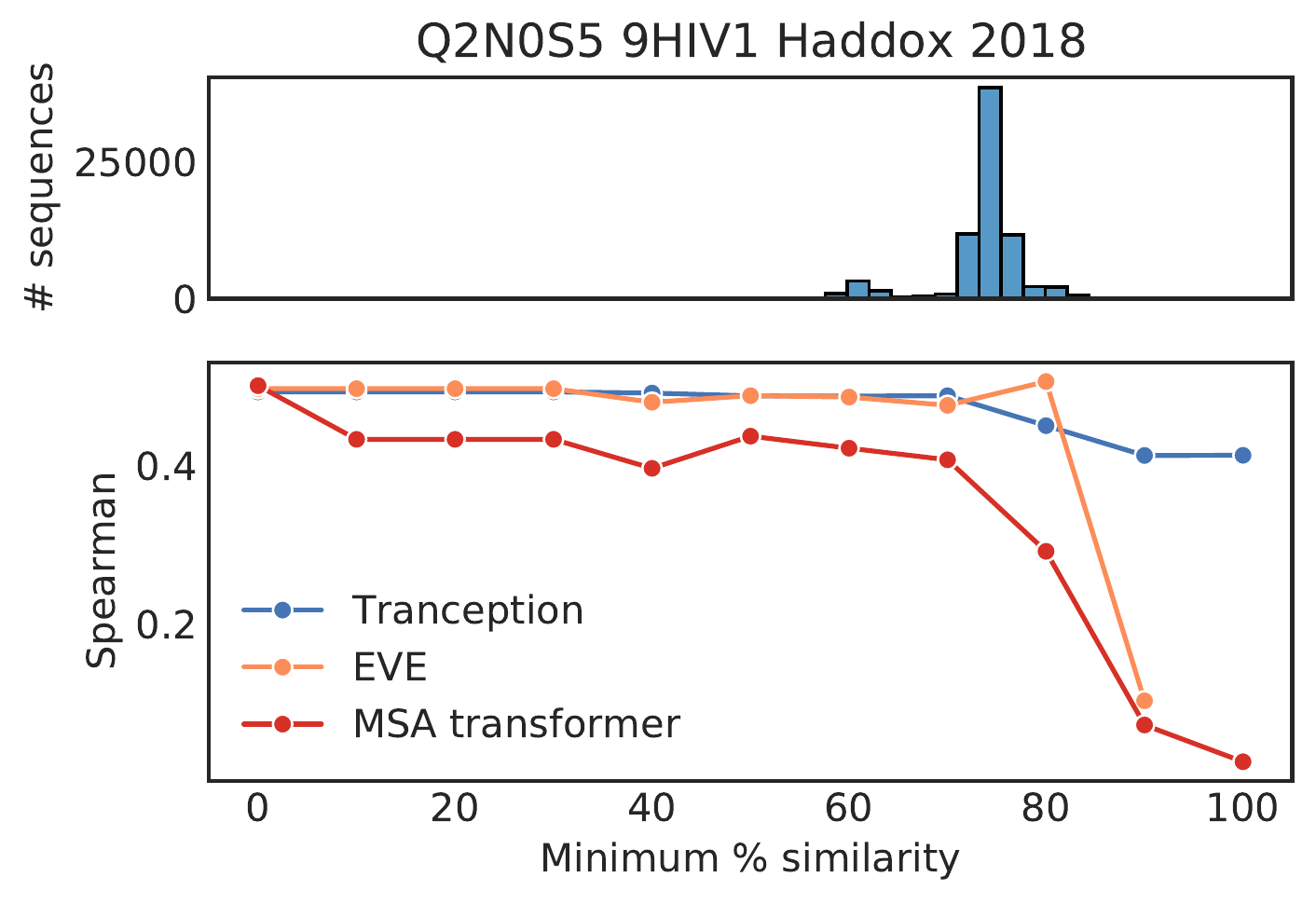}
    \hspace{0.01\linewidth}
    \includegraphics[ width = 0.3 \linewidth, keepaspectratio]{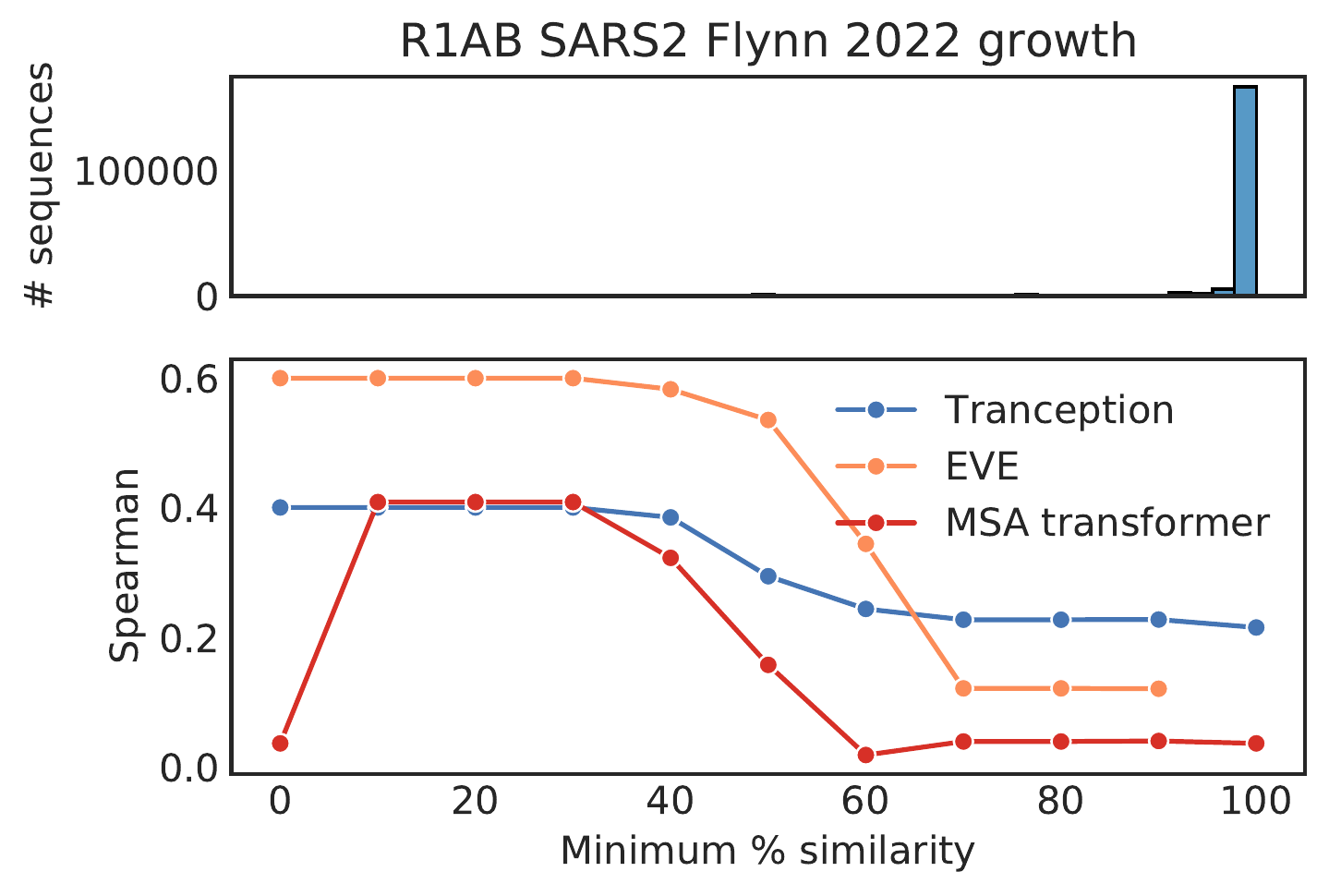}
    
    \vspace{0.03\textheight}
    
    \includegraphics[ width = 0.3 \linewidth, keepaspectratio]{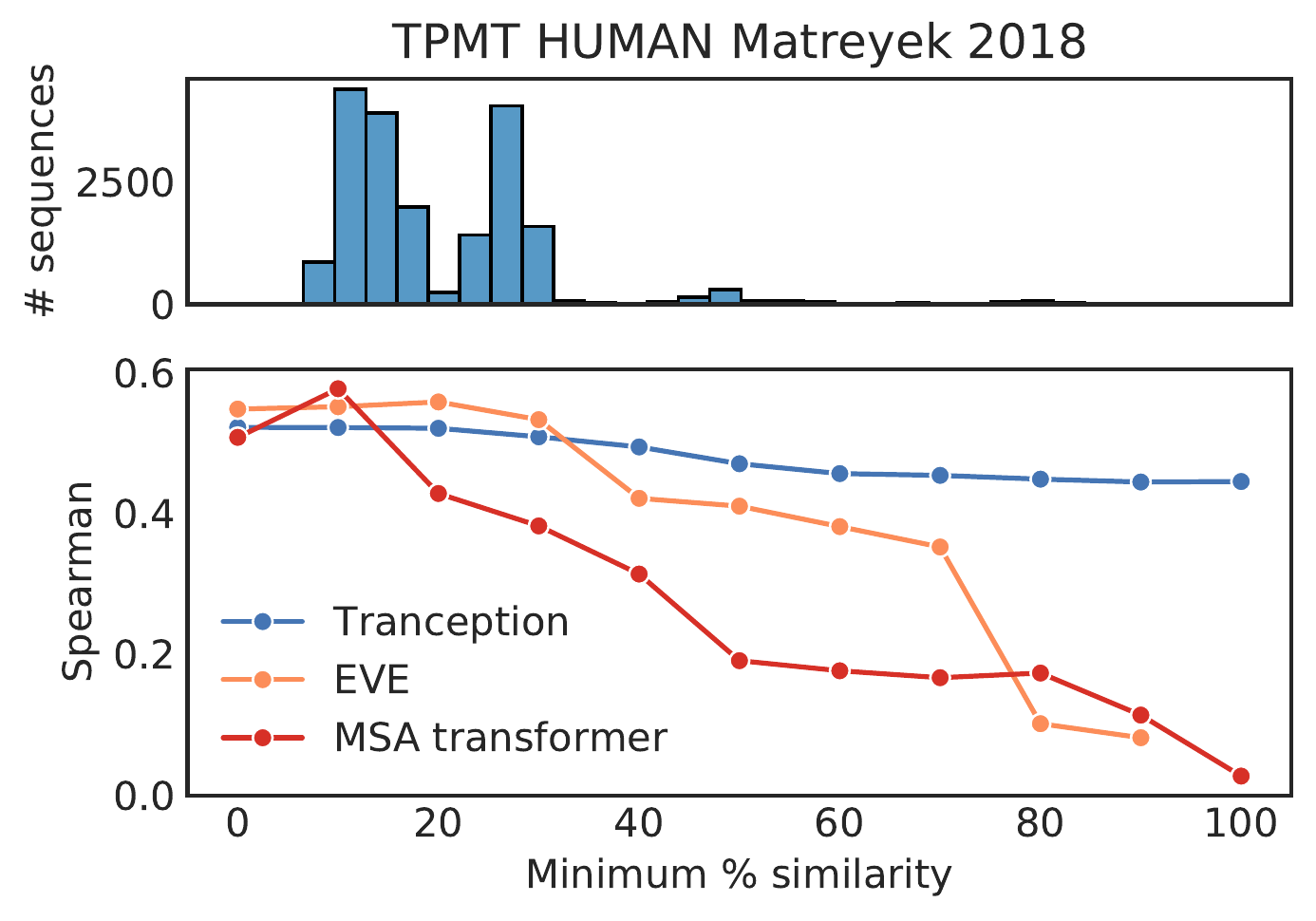}
    \hspace{0.01\linewidth}
    \includegraphics[ width = 0.3 \linewidth, keepaspectratio]{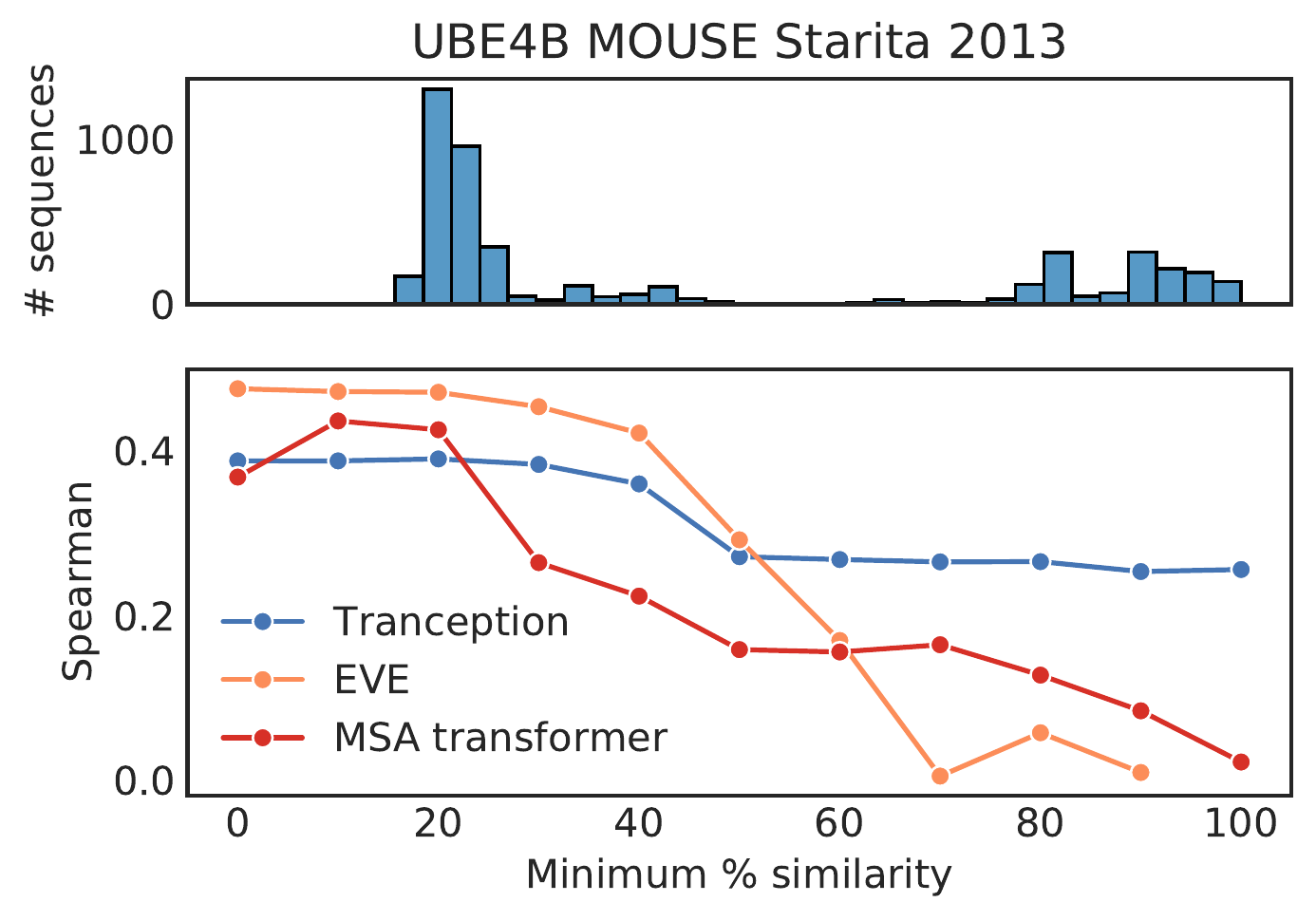}
    \hspace{0.01\linewidth}
    \includegraphics[ width = 0.3 \linewidth, keepaspectratio]{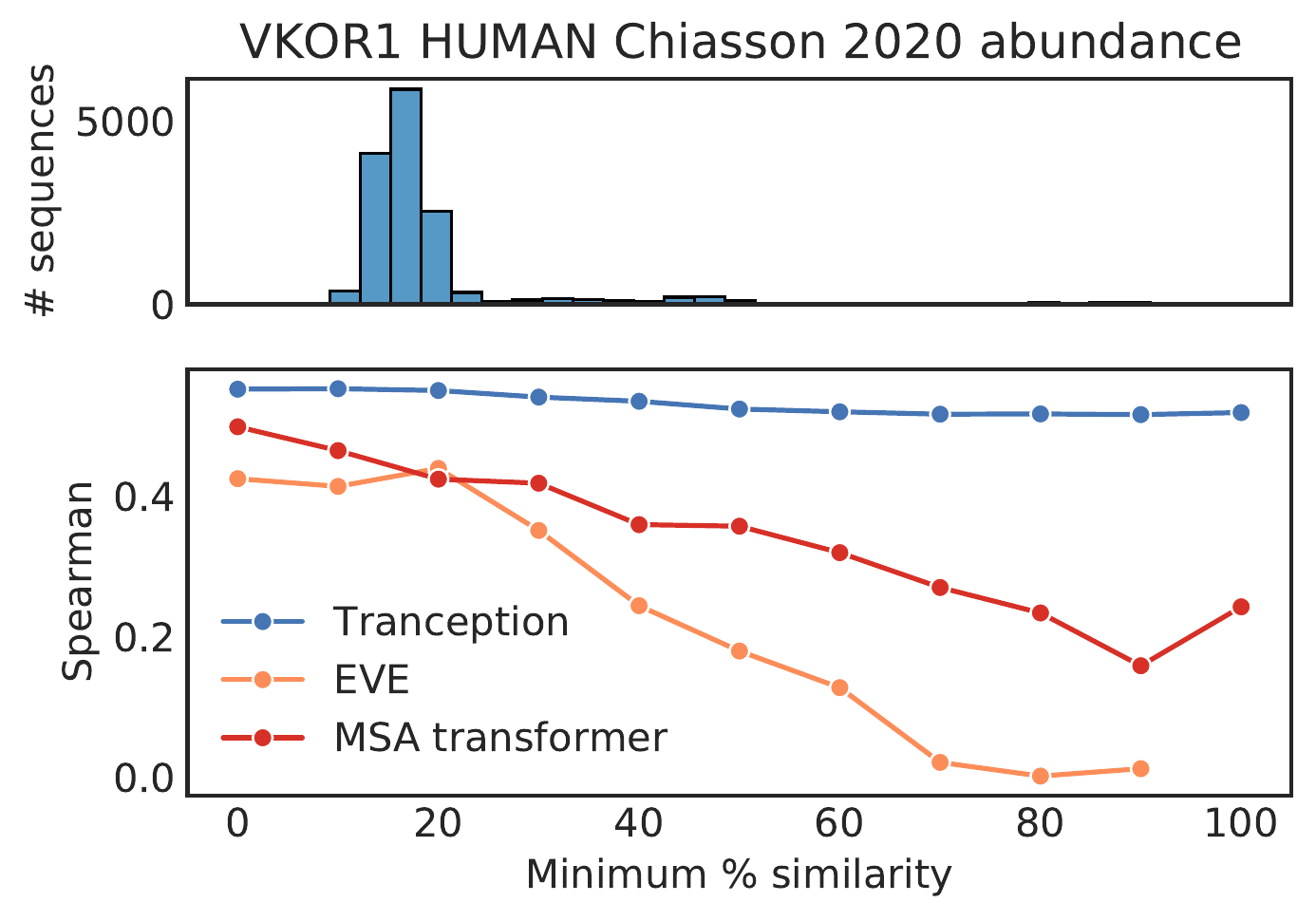}
     \caption{\textbf{Robustness to alignment depth for several DMS assays in the ProteinGym substitution benchmark}. We measure the Spearman's rank correlation between model score and experimental measurement for Tranception, EVE and MSA Transformer as we progressively exclude sequences in the corresponding MSAs based on their similarity to the sees sequence used to create the alignment.}
    \label{Appendix_fig:MSA_filtering_analysis}
\end{figure}

We report additional DMS level results for the MSA filtering analysis described in \S~\ref{Section: Discussion} in Fig.~\ref{Appendix_fig:MSA_filtering_analysis}. We observe that the performance of Tranception is less sensitive to MSA depth compared to EVE or MSA Transformer.

\subsection{Full protein Vs domain-specific models}
\label{Appendix:domain_specific_MSAs}

For all DMS assays we focused on full protein alignments. It may be possible to further increase the performance (and the coverage) of models leveraging MSAs by tailoring the creation of the MSA to known protein domains. For instance, we can obtain more diverse alignments for the RING and BRCT domains of BRCA1, thereby increasing the predictive performance of alignment-based models trained specifically on these alignments (Table~\ref{Appendix - Table: BRCA1 drill}). This also helps improve predictions made by Tranception with retrieval.

\begin{table}
    \centering
    \begin{tabular}{cccccc}
        \toprule
        Domain & Tranception & Tranception & Tranception &  EVE & EVE \\
               & (w/o retrieval) & (retrieval full MSA) & (retrieval domain MSA) & (full MSA) & (domain MSA) \\
        \midrule
       
       RING &  0.567 & 0.588 & \textbf{0.607} & 0.320 & 0.573 \\
       BRCT & 0.354 & 0.490 & 0.504 & N/A & \textbf{0.593} \\
    \bottomrule
    \end{tabular}
    \caption{\textbf{BRCA1 model performance summary by domain, as measured by Spearman's $\rho$ between model scores and experimental measurements} When using a full-protein alignment, EVE is unable to score mutations in the BRCT domain due to insufficient coverage in that region, as per the limitations discussed in Appendix~\ref{Appendix:baselines}}
    \label{Appendix - Table: BRCA1 drill}
\end{table}

\subsection{Model ensembling}
\label{Appendix: Ensembling}

Prior work from \citet{riesselman2018deep, Meier2021.07.09.450648} noted that additional performance gains on the fitness prediction task may be achieved in practice by ensembling several independently trained versions of the same model (eg., trained with different random seeds). We report the performance from several ensemble versions in Table~\ref{Appendix_table:single_archi_ensemble_performance}. Additionally, we revisit how ensembling has been approached so far for fitness prediction -- we propose that instead of ensembling several times the \emph{same} model architecture, we may obtain substantially higher performance in practice by ensembling \emph{different} yet complementary model architectures. We ensemble different pairs of models together and obtain the strongest results by combining Tranception with EVE (Table~\ref{Appendix_table:paired_archi_ensemble_performance}). 

\begin{table}
\parbox{.45\linewidth}{
    \centering
    \begin{tabular}{lcc}
        \toprule
        \multirow{2}{*}{\textbf{Model type}} &  \textbf{Single } &  \textbf{Ensemble of} \\
         &  \textbf{model} &  \textbf{5 models} \\
        \midrule
        Tranception w/o retrieval & 0.406 & - \\
        Tranception w/ retrieval & \textbf{0.451} & - \\
        ESM-1v & 0.371 & 0.401 \\
        MSA Transformer & 0.422 & 0.434 \\
        EVE & 0.448 & \textbf{0.452} \\
        \bottomrule
    \end{tabular}
    \caption{\textbf{Single-architecture ensemble analysis.} Tranception without retrieval (single seed) achieves higher average performance than the ESM-1v ensemble. Tranception with retrieval (single seed) achieves higher performance than the MSA Transformer ensemble. The ensemble of 5 EVE models does marginally better than a single Tranception model with retrieval. Performance is measured via Spearman's rank correlation between model scores and DMS measurements.}
    \label{Appendix_table:single_archi_ensemble_performance}
}
\hfill
\parbox{.45\linewidth}{
    \centering
    \begin{tabular}{lc}
        \toprule
        \textbf{Model pair ensembled} &  \textbf{Spearman} \\
        \midrule
        Tranception w/o retrieval only & 0.406 \\
        Tranception + ESM-1v & 0.427 \\
        Tranception + MSA Transformer & 0.449 \\ 
        Tranception + EVE & \textbf{0.473} \\
        \bottomrule
    \end{tabular}
    \caption{\textbf{Paired architecture ensemble analysis} In this analysis we always use Tranception without retrieval. We note however that ensembling Tranception with retrieval and EVE provides only marginally higher overall performance (0.475) suggesting that the retrieval inference may capture information analogous to that of alignment-based models like EVE, while the autoregressive inference provides different but complementary information.}
    \label{Appendix_table:paired_archi_ensemble_performance}
}
\end{table}

\section{ProteinGym curation}
\label{Appendix: Benchmark}

To build the ProteinGym benchmark, we initially collected a set of 137 deep mutational scanning assays. We then filtered out 43 of these assays based on the following criteria: Data had not been made publicly available (9), non-protein assays (UTR, tRNA, promoter, etc.; 7), synthetic proteins (3), insufficient number of measurements (3), outdated (ie., a more recent improved assay on the same protein and same property was found; 4), majority of data hitting experimental floor (6), low dynamic range (6), assay not relevant to fitness prediction (5). 

The final version of the ProteinGym consists of the experimental measurements of 94 deep mutational scanning experiments (87 substitutions assays and 7 indels assays) from the following 77 publications: \cite{adkar_protein_2012, jones_structural_2020, kozek_high-throughput_2020, firnberg_comprehensive_2014, jia_massively_2021, chan_correlation_2017, wu_functional_2015, mavor_determination_2016, ahler_combined_2019, newberry_robust_2020, fernandes_functional_2016, mighell_saturation_2018, roscoe_systematic_2014, kotler_systematic_2018, rockah-shmuel_systematic_2015, giacomelli_mutational_2018, melamed_deep_2013, melnikov_comprehensive_2014, brenan_phenotypic_2016, suiter_massively_2020, chen_comprehensive_2020, tripathi_molecular_2016, pokusaeva_experimental_2019, aakre_evolving_2015, haddox_experimental_2016, soh_comprehensive_2019, deng_deep_2012, weile_framework_2017, klesmith_comprehensive_2015, doud_accurate_2016, wu_high-throughput_2015, findlay_accurate_2018, stiffler_evolvability_2015, faure2022mapping, jacquier_capturing_2013, kennouche_deep_2019, sourisseau_deep_2019, wrenbeck_single-mutation_2017, dandage_differential_2018, seuma_genetic_2021, haddox_mapping_2018, doud_site-specific_2015, mishra_systematic_2016, flynn_comprehensive_2020, amorosi_massively_2021, mclaughlin_jr_spatial_2012, nutschel_systematically_2020, kitzman_massively_2015, kelsic_rna_2016, lee_deep_2018, mattenberger_globally_2021, matreyek_multiplex_2018, thompson_altered_2020, romero_dissecting_2015, qi_quantitative_2014, roscoe_analyses_2013, bandaru_deconstruction_2017, young_deep_2021, glazer_deep_2020, chiasson_multiplexed_2020, olson_comprehensive_2014, bridgford_novel_2020, starita_activity-enhancing_2013, starr_deep_2020, araya_fundamental_2012, davidi2020highly, russ2020evolution, gonzalez2019fitness, sinai2021generative, bolognesi2019mutational, duenas2016saturation, jiang2016balance, matreyek2021integrating, sarkisyan2016local, staller2018high}. 

We preprocessed the raw DMS assays as indicated in \S~\ref{Appendix:performance_reporting}. The ProteinGym benchmarks are made publicly available (both raw and processed assay data) on our \href{https://github.com/OATML-Markslab/Tranception}{GitHub repository}.


\end{document}